\documentclass[preprint,12pt]{elsarticle}

\usepackage{amssymb}
\usepackage{amsmath}
\usepackage{booktabs}
\usepackage{subcaption}
\usepackage{multirow}
\usepackage{xcolor}
\newcommand{\Adam}{{\sc Adam}}

\journal{Knowledge-Based Systems}

\begin{document}

\begin{frontmatter}



\title{MoTE: Mixture of Task-specific Experts for Pre-Trained Model-Based Class-incremental Learning}


\author[label1,label2]{Linjie Li}
\author[label1,label2]{Zhenyu Wu \corref{cor1}}
\author[label1,label2]{Yang Ji}
\cortext[cor1]{Corresponding author, E-mail: shower0512@bupt.edu.cn}
\affiliation[label1]{organization={School of Information and Communication Engineering, Beijing University of Posts and Telecommunications},%Department and Organization
            city={Beijing},
            postcode={100876}, 
            country={China}}
\affiliation[label2]{organization={Engineering Research Center for Information Network, Ministry of Education},%Department and Organization
            city={Beijing},
            postcode={100876}, 
            country={China}}
%% Abstract
\begin{abstract}

Class-incremental learning (CIL) requires deep learning models to continuously acquire new knowledge from streaming data while preserving previously learned information. Recently, CIL based on pre-trained models (PTMs) has achieved remarkable success. However, prompt-based approaches suffer from prompt overwriting, while adapter-based methods face challenges such as dimensional misalignment between tasks. While the idea of expert fusion in Mixture of Experts (MoE) can help address dimensional inconsistency, both expert and routing parameters are prone to being overwritten in dynamic environments, making MoE challenging to apply directly in CIL. To tackle these issues, we propose a mixture of task-specific experts (MoTE) framework that effectively mitigates the miscalibration caused by inconsistent output dimensions across tasks. Inspired by the weighted feature fusion and sparse activation mechanisms in MoE, we introduce task-aware expert filtering and reliable expert joint inference during the inference phase, mimicking the behavior of routing layers without inducing catastrophic forgetting. Extensive experiments demonstrate the superiority of our method without requiring an exemplar set. Furthermore, the number of tasks in MoTE scales linearly with the number of adapters. Building on this, we further explore the trade-off between adapter expansion and model performance and propose the Adapter-Limited MoTE. The code is available at https://github.com/Frank-lilinjie/MoTE.

\end{abstract}

% \includegraphics[width=\textwidth]{overveiw.png}

% \begin{highlights}
% \item Injects task-specific knowledge into frozen models via lightweight adapters.
% \item Adopts MoE-inspired strategy to address challenges in adapter-based CIL.
% \item Proposes MoTE, a more forgetfulness-resistant method than traditional MoE.
% \item Demonstrates MoTE's superiority, robustness, and generalizability across benchmarks.
% \item Adapter-Limited MoTE explores the trade-off between adapter count and performance.
% \end{highlights}

%% Keywords

\end{frontmatter}



\section{Introduction}
%% Labels are used to cross-reference an item using \ref command.

The significant advancements in deep learning have led to the widespread application of deep neural networks in the real world~\cite{IAI_1,KBS_1,IAI_2,KBS_2,imagenet}. However, the challenge posed by technological development is that new data is being generated at an exponentially increasing rate. Additionally, in the real world, data exists in a streaming format~\cite{stream_1,stream_2}. The costs associated with training models from scratch in terms of time and computational resources are becoming increasingly prohibitive. Therefore, deep learning models must have the capability for continual learning. Class-incremental learning (CIL) is a specific paradigm of continual learning, where the objective is to allow models to learn new tasks while progressively accumulating and retaining knowledge over time~\cite{CILsurvey,Continualsurvey}. However, when learning new tasks, the representations of old tasks may inevitably be disrupted, resulting in a sharp decline in performance on prior tasks—which is known as catastrophic forgetting~\cite{catastrophic,cataph_2}. To mitigate catastrophic forgetting, many solutions have been proposed, including data replay~\cite{icarl,EEIL}, knowledge distillation~\cite{lwf,podnet}, dynamic networks~\cite{DER,Foster,memo}, and parameter regularization~\cite{ewc}.

\begin{figure}[t]
\centering
\includegraphics[width=0.8\columnwidth]{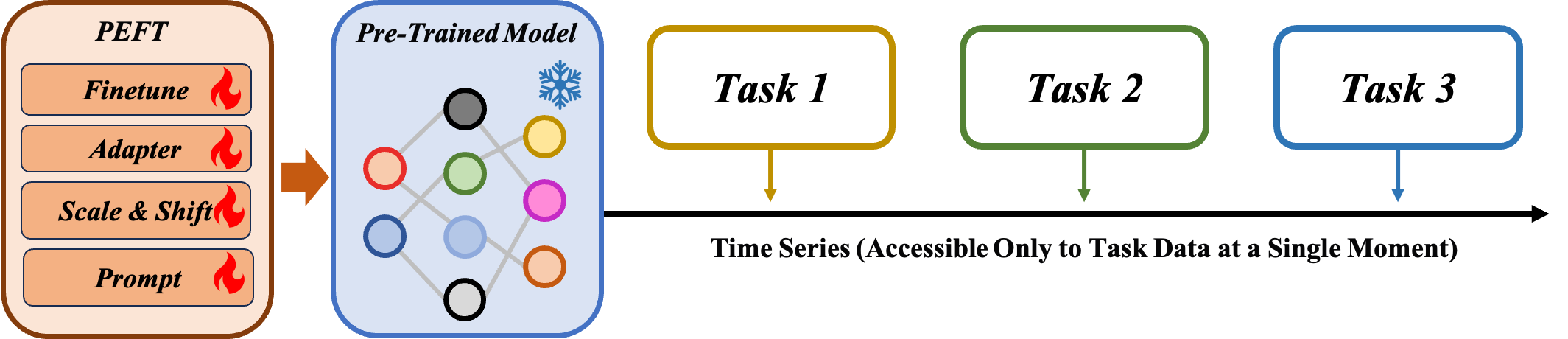}
\caption{Overview of PTM-based CIL Algorithms, these algorithms freeze the pre-trained model and leverage parameter-efficient fine-tuning (PEFT) techniques, including \textit{Finetune, Adapter, Scale $\&$ Shift}, and \textit{Prompt}, to adapt to continuously incoming downstream tasks with minimal additional model parameters. The model learns downstream streaming tasks sequentially, where only the current task data is accessible at each time step.}
\label{fig: CLmeaning}
\end{figure}

\begin{figure}[t]
	\centering
	\begin{subfigure}{0.635\linewidth}
		\includegraphics[width=1\columnwidth]{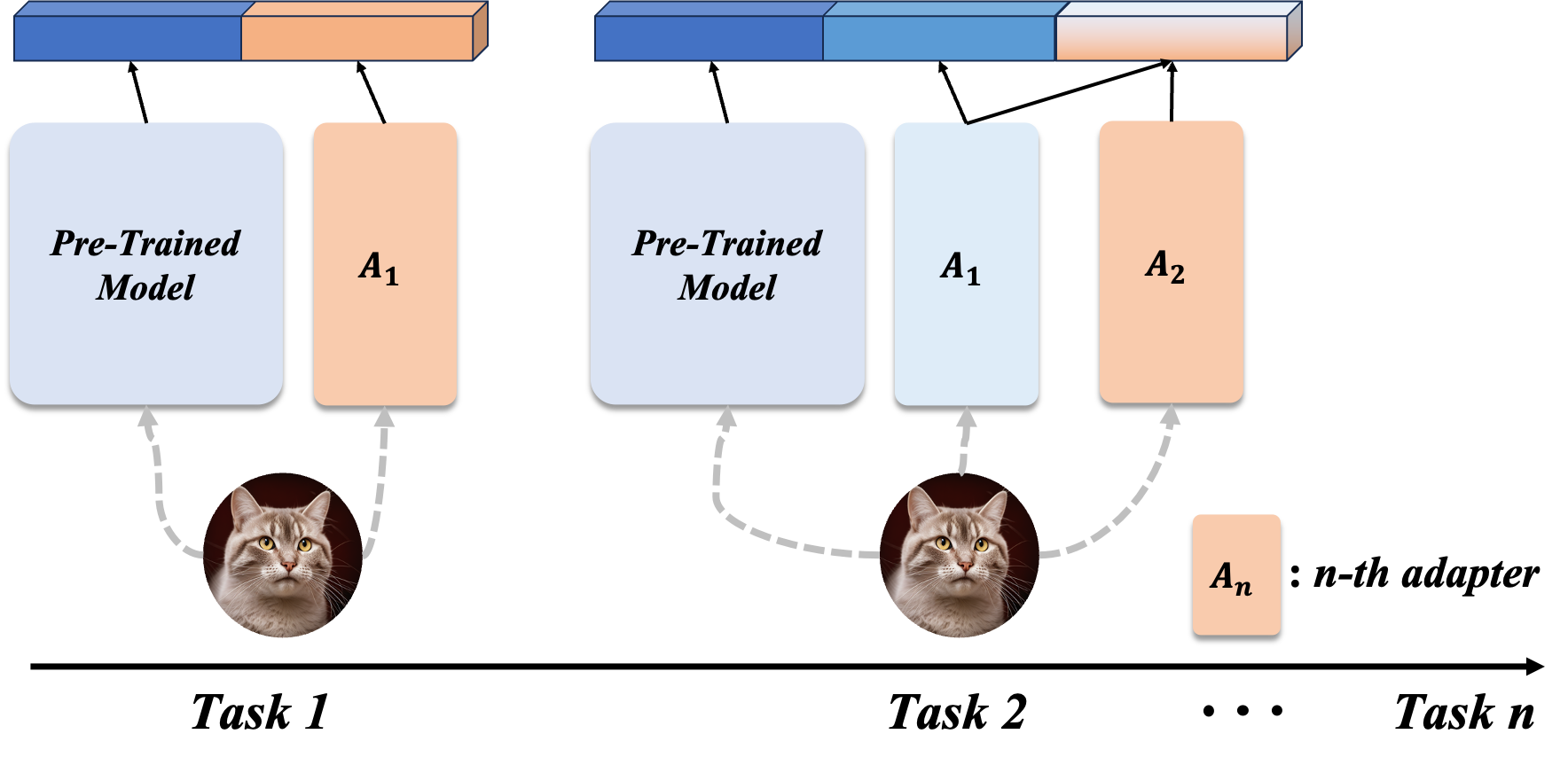}
            \caption{}
		\label{fig:mofe}
	\end{subfigure}
	\hfill
	\begin{subfigure}{0.345\linewidth}
		\includegraphics[width=1\linewidth]{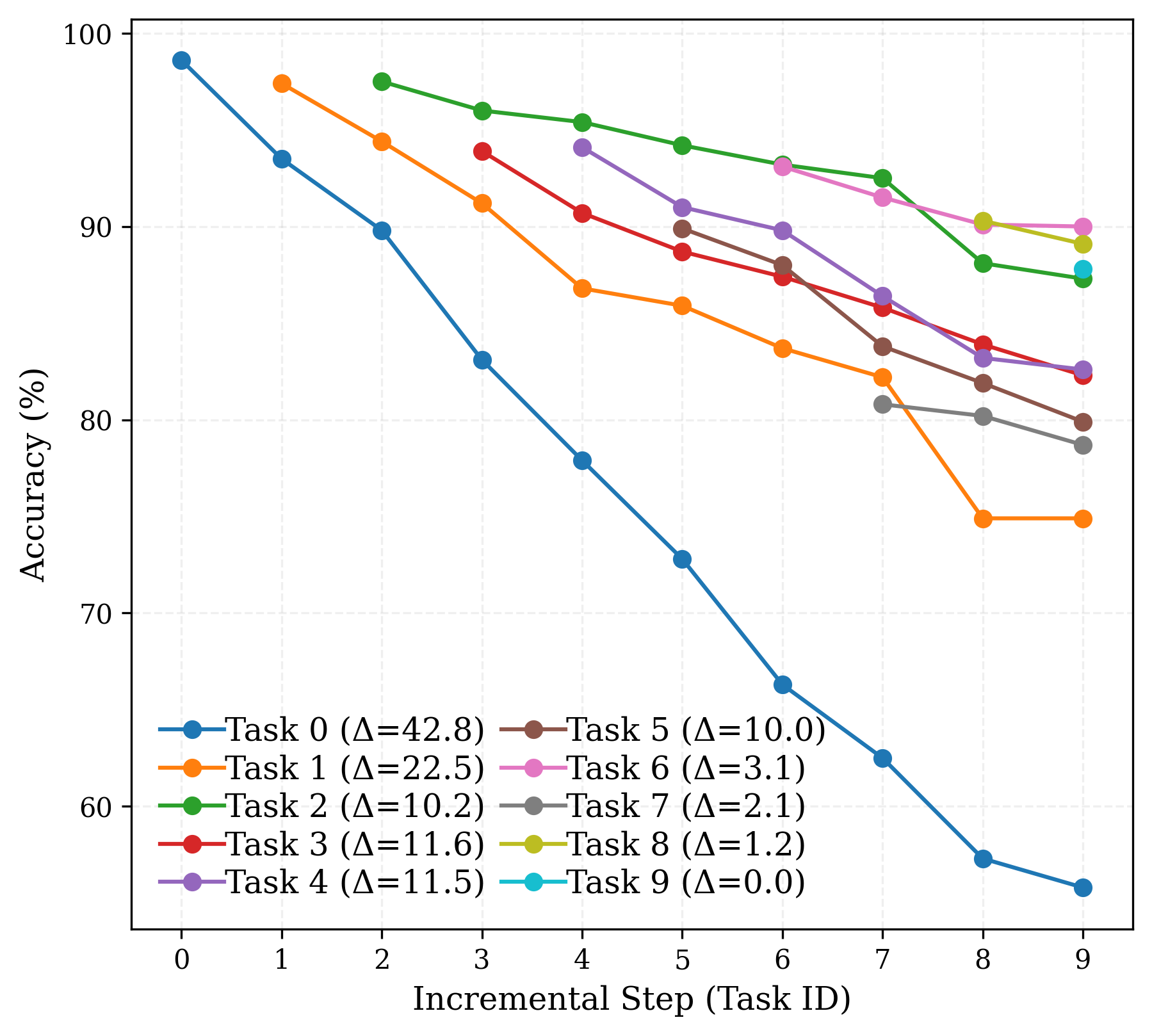}
            \caption{}
		\label{fig:gap0}
	\end{subfigure}
    \caption{Illustration of research motivation and preliminary experiment. \textbf{Left}: Adapter expansion based on the concept of dynamic networks. When learning a new task, previous adapters are frozen while a new adapter is trained, leading to dimensional expansion. Previous adapters influence the learning of new task representations, and the new adapter affects the inference of old task samples. \textbf{Right}: Accuracy decay curves for each task across incremental steps. At each step, the network grows as new adapters are added. The legend indicates the task ID and the accuracy drop from the initial to the final step. (CIFAR B0-Inc10).}
\end{figure}

Recently, driven by the flourishing development of pre-trained foundational models, CIL algorithms based on pre-trained models (PTMs) have demonstrated exceptional performance~\cite{Continualsurvey,Pilot,ADAM}. PTMs are trained on large-scale datasets using substantial computational resources, endowing powerful representational capabilities~\cite{pretrain}, including zero-shot learning abilities. To ensure that the representations of PTMs are not compromised during continual learning training, most current research begins by freezing the PTM backbone and adding trainable modules with parameter efficient fine tuning (PEFT) techniques to adapt to new tasks, as shown in Fig~\ref{fig: CLmeaning}. Prompt-based algorithms have gained popularity~\cite{l2p,dualprompt,Hide,codaprompt}. In brief, these algorithms establish a prompt pool and initialize multiple pairs of keys and prompts, selecting instance-specific prompts for training. Compared to traditional continual learning methods, this approach has shown superior performance. However, since the number of prompts in the prompt pool is fixed, fitting new tasks inevitably leads to the rewriting of prompts for old tasks, resulting in forgetting. Compared to prompts, the lightweight adapter is embedded in the model structure, providing guiding representations. However, suppose we follow the dynamic network approach to extend adapters. In that case, we assign a new adapter for each task during training while freezing the adapters from previous tasks. The outputs from multiple adapters are then concatenated, as shown in Fig.~\ref{fig:mofe}. Adding new adapters changes the model's dimensionality, which alters the representation of the old task samples, leading to catastrophic forgetting and a decline in model performance, as illustrated in Fig.~\ref{fig:gap0}. A seemingly straightforward approach is to use a Mixture of Experts (MoE)~\cite{MoE}. However, in dynamic training environments, the routing layer and expert parameters are prone to being overwritten or disrupted. Some studies replace the linear routing layer by using autoencoders and reconstruction thresholds to infer the task identity of a test sample. However, introducing autoencoders significantly increases storage and training costs. Moreover, the reconstruction threshold is manually defined and requires extensive experimentation to determine a suitable range. This threshold is also sensitive to changes in data distribution, necessitating recalibration across different settings, which limits the generalizability of such methods~\cite{moe_gating,moe_clip}. Therefore, effectively applying MoE to CIL remains a non-trivial and challenging task.

To address the above issues, we propose a mixture of task-specific experts (MoTE) approach that effectively mitigates the miscalibration problem caused by inconsistent output dimensions across tasks. Inspired by the weighted feature fusion and sparse activation mechanisms in MoE~\cite{MoE2017}, our method introduces expert filtering and weighted multi-expert joint inference during the inference stage. This design emulates the behavior of a routing mechanism while avoiding the catastrophic forgetting commonly associated with dynamic updates. Specifically, we keep the pre-trained model frozen during the training phase and train a lightweight adapter for each task independently. This allows each adapter to fully capture task-specific knowledge without interference from other tasks, effectively functioning as a task-specific expert. After training, we extract features from all samples within each class and compute their mean to obtain class prototypes, which serve as anchors for classification during inference ~\cite{ADAM,EASE,prototype}. We first perform expert filtering based on each expert’s prediction at inference time. If an expert predicts a class outside its designated task scope, it is labeled as “unreliable” and excluded. Conversely, if the prediction falls within its task’s label space, it is marked as a “reliable” expert and retained. This mechanism enforces sparse expert activation during inference. For easily distinguishable samples, typically, only one reliable expert is activated. However, when multiple reliable experts are retained, it indicates ambiguity in task affiliation. In such cases, we perform a weighted multi-experts joint inference by combining the outputs of these experts. The weights are adaptively determined using both confidence and self-confidence scores(SCS)~\cite{SCS}, giving more influence to the most relevant experts and enhancing the model’s robustness. Our approach eliminates the need to train an explicit routing module, thereby avoiding parameter overwriting issues caused by distribution shifts during continual training. Moreover, our weighting strategy is fully adaptive, relying only on expert predictions and introducing no additional hyperparameters, which improves the model’s generalizability.

We conducted experiments on several widely used benchmark datasets, including CIFAR100~\cite{CIFAR}, CUB200~\cite{CUB}, ImageNet-A~\cite{ImageNetA}, ImageNet-R~\cite{ImageNetR}, and VTAB~\cite{VTAB}. Extensive performance comparisons and ablation studies validate the effectiveness and robustness of MoTE. Additionally, we conduct in-depth analyses, including the comparison of global and adaptive Scaling Factors, the comparison of inference time, and visualizations. Notably, our proposed method demonstrates superior performance on multiple benchmarks while achieving an inference speed that is almost 30$\%$ faster than the state-of-the-art (SOTA) method.

Moreover, in MoTE, the number of adapters grows linearly with the number of tasks. However, as shown in our ablation experiments in Sect.~\ref{Sec:ablation_limit}, this is a suboptimal strategy. If adapters could be adaptively expanded based on task complexity, the algorithm could become more lightweight and efficient.  To address this, we propose Adapter-Limited MoTE, which aims to restrict the number of adapters while evaluating the model’s incremental learning capability. We hope this study provides valuable insights and inspiration for future research.

The contributions of this paper are as follows:
\begin{itemize}

\item We propose a novel exemplar-free mixture of task-specific experts(MoTE) framework that effectively addresses the miscalibration caused by inconsistent output dimensions across tasks. Inspired by the weighted feature fusion and sparse activation mechanisms in Mixture of Experts (MoE), our method introduces task-aware expert filtering and reliable expert joint inference during the inference phase, simulating the behavior of a routing layer while avoiding catastrophic forgetting.

\item Furthermore, we explore the relationship between the number of tasks and the extent of adapter expansion. Building on MoTE, we develop Adapter-Limited MoTE and conduct comparative experiments to evaluate the impact of limiting the number of adapters. We hope this exploration provides valuable insights for future research.

\item Comprehensive experiments and in-depth analysis. We conduct extensive experiments to validate the effectiveness of MoTE. The results demonstrate that MoTE achieves state-of-the-art performance on standard benchmarks while being 30$\%$ faster in inference compared to the SOTA method.

\end{itemize}

%% Use \subsection commands to start a subsection.
\section{Related Work}
\subsection{Class-incremental Learning}

Class-incremental Learning (CIL) requires deep learning systems to acquire knowledge of new tasks while retaining knowledge of previous ones. The greatest challenge in this process is the notorious issue of catastrophic forgetting~\cite{catastrophic,CILsurvey,Continualsurvey}. Traditional CIL algorithms can be broadly categorized into several types. \textbf{Replay-based} methods preserve knowledge of old tasks by retaining a small subset of previous data and replaying it during the learning of new tasks. Rebuffi et al. were the first to propose introducing a small number of exemplars from old classes in class-incremental learning~\cite{icarl}. They also introduced a method called "\textit{herding}" to select these exemplars. Liu et al. introduced a bi-level optimization framework that distills new class data into exemplars before discarding it~\cite{liu2020mnemonics}. Wang et al. aimed to trade off between the quality and quantity of exemplars by image compression using the JPEG algorithm~\cite{wang2022memory}. Luo et al. proposed a class-incremental masking approach that downsamples only the non-discriminative pixels in the images~\cite{liu2023CIM}. \textbf{Dynamic-parameter-based} methods learn new knowledge by expanding the network capacity while preserving knowledge of old tasks by freezing the previous network parameters. Yan et al. proposed extending the network backbone when encountering a new task and subsequently aggregating the new features at the classifier level to handle multiple tasks effectively~\cite{DER}. Zhou et al. tackled the issue of excessive memory overhead by decoupling the intermediate network layers and dynamically expanding the network, which mitigates the memory budget challenges associated with deep network extensions~\cite{memo}. In contrast to architecture-based expansion, Li et al. further explored parameter-based expansion in the context of long-tailed class-incremental learning~\cite{li2024tae}. \textbf{Regularization-based} methods balance the knowledge between new and old classes by adjusting the model’s output logits~\cite{lwf,wu2019large}, intermediate features~\cite{hou2019learning,dhar2019learning}, or relationships between classes~\cite{gao2022r,dong2021few} through penalty terms or knowledge distillation.

\subsection{Pre-Trained Models and Parameter-Efficient Fine-Tuning}
Researchers train large-scale models with billions of parameters using massive datasets and apply them to downstream tasks. These pre-trained models possess zero-shot capabilities, and transferring them to downstream tasks can significantly improve performance. While pre-trained models can generate generalized features and adapt to various domains, their performance often falls short compared to domain-specific expert models. Moreover, fully fine-tuning pre-trained models for downstream tasks not only requires substantial computational resources but can also negatively impact their generalization ability. Consequently, one of the most prominent paradigms in artificial intelligence today is the development of efficient techniques to fine-tune pre-trained models by adjusting or embedding a small number of trainable parameters to adapt them to downstream tasks better. Visual Prompt Tuning (VPT)~\cite{vpt} introduces tunable prefix tokens~\cite{prefix} that are appended to either the input or hidden layers. Low-Rank Adaptation (LoRA)~\cite{lora} leverages low-rank matrices to approximate parameter updates, while AdaptFormer~\cite{adaptformer} incorporates additional adapter modules with downsize and upsize projections. Scale and Shift deep Features (SSF) ~\cite{ssf} optimizes model tuning by addressing scaling and shifting operations.

\subsection{Class-incremental Learning with Pre-Trained Models}

\begin{figure}[t]
\centering
\includegraphics[width=1.0\columnwidth]{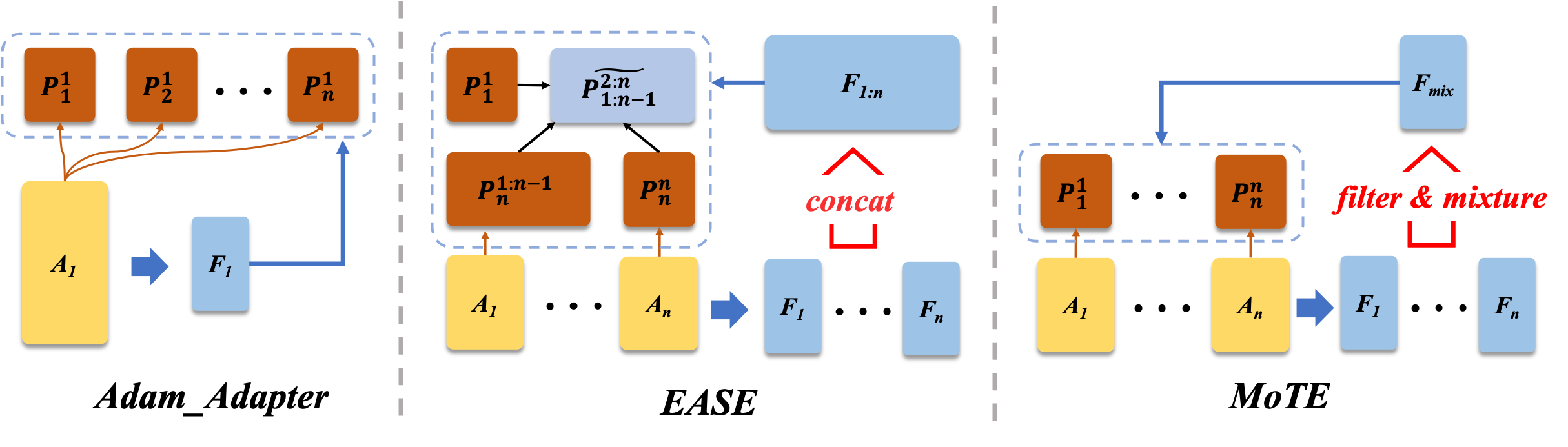}
\caption{Illustrative comparison of related methods. $P_{a}^{b}$ represents the class prototype generated for dataset $a$ by the $b-th$ expert and $\tilde{P}$ represents the pseudo-class prototypes that require additional computation.; $A_{n}$ represents $n-th$ adapter; $F_{n}$ represents the features generated by the $ n-th$ adapter.}
\label{fig:compare}
\end{figure}

With the rise of research on PTM, PTM-based CIL has also gained popularity~\cite{Continualsurvey,Pilot,PTMsurveyCIL}. The goal is to continually adapt PTM to streaming downstream tasks while preventing the forgetting of patterns from previous tasks. Currently, PTM-based CIL methods can be categorized into two types. \textbf{Prompt-based} methods involve building a vision prompt pool and selecting sample-specific prompts from the pool to incorporate into the model for adapting to downstream tasks. Wang et al. integrated visual prompt tuning into class-incremental learning (CIL), utilizing a pre-trained Vision Transformer and establishing a prompt pool to select instance-specific prompts~\cite{l2p}. Building upon this, they proposed two types of prompts: general prompts and expert prompts~\cite{dualprompt}. Smith et al. enhanced the selection process by incorporating an attention mechanism to improve prompt selection~\cite{codaprompt}. Wang et al. conducted a theoretical analysis of the continual learning objective in the context of pre-training and decomposed it into hierarchical components: within-task prediction, task-identity inference, and task-adaptive prediction~\cite{Hide}. \textbf{Model-based} methods, on the other hand, adjust a few parameters of the model or embed a small number of additional parameters through adaptation modules. Zhang et al. updated the entire backbone with a reduced learning rate and further preserved pre-trained representations using dedicated covariance matrices~\cite{slca}. McDonnell et al. projected pre-trained representations into a high-dimensional space and retained them through a shared covariance matrix~\cite{ranpac}. Zhou et al. generate pseudo-class prototypes for old tasks on new adapters by calculating the similarity between old and new task class prototypes~\cite{EASE}. During inference, results from all adapters are combined and compared against the full set of class prototypes. However, the computation of pseudo-class prototypes incurs additional overhead. Additionally, Zhou et al. revisited the use of pre-trained models in CIL, arguing that the core factor of CIL is the model's ability to transfer knowledge to downstream tasks~\cite{ADAM}. \Adam\_Adapter~\cite{ADAM} integrates adapters into the pre-trained model and performs training only during the first round, keeping both the model and adapters frozen. It directly generates class prototypes for all classes without additional updates. MoTE generates independent class prototypes for each task. During inference, the expert evaluation mechanism determines the weight distribution for the expert combination. The combined results from multiple experts are then compared against all class prototypes. The three methods are compared in Fig.~\ref{fig:compare}.

\subsection{Mixture of Experts}
Mixture of Experts (MoE) is a classic approach that combines specialized sub-models (experts) through a gating network to handle diverse data distributions~\cite{MoE}. In traditional MoE, a trainable gating network dynamically assigns weights to experts based on input features, producing a weighted combination of expert representations. To improve computational efficiency, Shazeer et al. introduced sparse activation, where only a subset of experts (e.g., 1–2 experts) is activated per sample~\cite{MoE2017}. Riquelme et al. further applied this sparse MoE paradigm to computer vision tasks~\cite{V-MoE}. Inspired by these works, we introduce adapters as task-specific experts in our incremental learning framework, leveraging newly added experts and expert fusion to facilitate CIL. However, applying traditional MoE to CIL presents several challenges. First, conventional MoE relies on a \textbf{data-driven} gating mechanism, but in the CIL setting, data is sequentially streamed, and previously seen data is inaccessible. This necessitates additional strategies to mitigate catastrophic forgetting in the gating network. Second, during training, feature interference between new and old experts can degrade performance, making it difficult to effectively integrate expert representations via the gating mechanism. Recent studies have explored the use of Mixture of Experts (MoE) in continual learning. For instance, Aljundi et al. introduced autoencoders to mitigate routing bias by using reconstruction thresholds to determine which expert should handle a given inference sample~\cite{moe_gating}. Similarly, Yu et al. adopted this idea and combined MoE with CLIP for continual learning~\cite{moe_clip}. However, employing autoencoders incurs additional storage and training costs. Moreover, the reconstruction threshold must be manually set, often requiring extensive experimentation to identify a reasonable range, which can vary significantly depending on the data distribution, thus limiting the generalizability of this approach. These challenges highlight that directly applying MoE to class-incremental learning is far from straightforward. To address these challenges, we propose \textbf{Task-driven} MoTE (Mixture of Task-Specific Experts), where each task is explicitly assigned a dedicated adapter, ensuring independent learning and preventing expert interference. To mitigate gating network forgetting, we introduce an expert filtering mechanism based on predefined task scopes and expert output features, ensuring that only task-relevant experts participate in inference, analogous to sparse activation in MoE. Consequently, MoTE effectively tackles the unique challenges of CIL by incorporating task-aware expert filtering and confidence-weighted dense fusion, preserving task-specific knowledge while enabling adaptive expert integration for efficient incremental inference.

\section{Preliminaries}
\subsection{Problem Formulation}
Class incremental learning (CIL) aims to build a unified classifier from streaming data. In PTM-based CIL, the focus is primarily on exemplar-free CIL, where only the data from the current task is accessible when learning new tasks. For formalization, we denote the sequentially presented datasets for $t$ training tasks as $\left \{{\mathcal{D}^{1}}, {\mathcal{D}^{2}} \cdots {\mathcal{D}^{t}}\right \} $, where each $\mathcal{D}^{t}$ consists of $n_{t}$ pairs $(\mathbf{x}_{i}, \mathbf{y}_{i})$, \((\mathbf{x}, y)\) represents a data pair, where \(\mathbf{x}\) denotes the input sample and \(y\) corresponds to its associated label. In this paper, we adhere to the problem settings of previous CIL research, and cases of class overlap or multi-label co-occurrence are not within the scope of our research~\cite{CILsurvey, PTMsurveyCIL, Continualsurvey}. Therefore, the objective of CIL is to minimize the expected risk across all test datasets:
\begin{equation}
\label{Expected_Risk}
f^{\ast} = \underset{f\in \mathcal{H} }{\mathrm{argmin}} \mathbb{E} _{(\mathbf{x},\mathbf{y} )\sim \mathcal{D}^{1}\cup \cdots \mathcal{D}^{t}}\mathbb{I}(\mathbf{y}\ne f(\mathbf{x})) 
\end{equation}
Here, $f^{\ast}$ represents the optimal model, $\mathcal{H}$ is the hypothesis space, and $\mathbb{I}(\cdot)$ denotes the indicator function. $\mathcal{D}^{t}$ represents the data distribution of the task $t$, $\mathbf{y}$ represents a one-hot vector. 

\subsection{Pre-Trained Model and Adapter Techniques}
This paper decouples the model into feature extraction and linear classification layers. For the feature extraction layer, in standard pre-trained continual learning algorithms~\cite{codaprompt,l2p,dualprompt}, we initialize model parameters using a Visual Transformers (ViT) ~\cite{ViT} as the pre-trained model, denoted as $f(x)$. The ViT structure consists of multiple consecutive multi-head attention (MSA) and feed-forward network (FFN) layers. Let $h_{l_{in}}$ represent the input at layer $l$, $h_{l_{msa}}$ represent the output of MSA layers, and $h_{l_{out}}$  represent the output of the block. For each input $h_{in}$ at a given layer, it is split into query $h_{Q}$, key $h_{K}$, and value $h_{V}$. $h_{msa}$ can then be computed as follows:
\begin{equation}
h_{msa} = \mathrm{MSA} (h_{Q},h_{K},h_{V})=Concat(h_{1},...,h_{m})W_{O}
\end{equation}
\begin{equation}
h_{i} = \mathrm{Attn} (h_{Q}W_{Q,i},h_{K}W_{K,i},h_{V}W_{V,i}),i\in [m]
\end{equation}
Here, $W_{O},W_{Q,i},W_{K,i},W_{V,i}$ are projection matrices, and $m$ is the number of attention heads. In the ViT, $h_{Q}=h_{K}=h_{V}=h_{in}$. The concatenation (Concat) and attention (Attn) operations are defined as in their sources~\cite{Transformer,ViT}, the computation formula for Attention is:
\begin{equation}
h_{i} = softmax(\frac{(h_{Q}W_{Q,i})(h_{K}W_{K,i})^{T}}{\sqrt{d_{k}} } )(h_{V}W_{V,j})
\end{equation}
Here, $d_{k}$ represents the dimension of the key vector.
\begin{figure}[t]
\centering
\includegraphics[width=0.6\columnwidth]{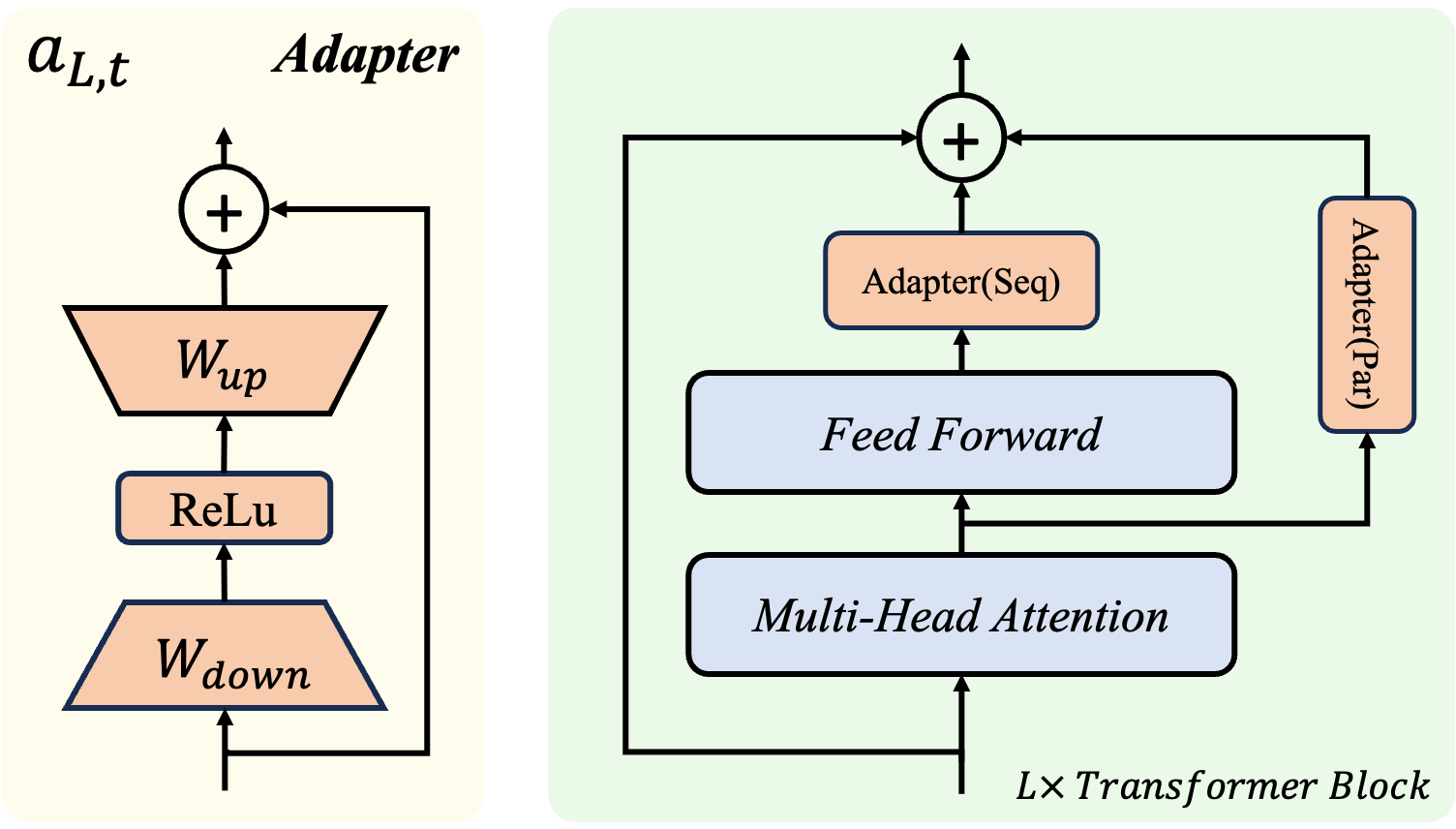}
\caption{Structure of the adapter and two embedding types: sequential(Seq) and parallel(Par).}
\label{fig:adapter}
\end{figure}

Adapters \cite{adapter} introduce lightweight modules in the blocks of the backbone. Each module typically consists of a down-projection matrix $W_{down}\in \mathbb{R}^{d\times r}$,  which reduces the dimensionality of the hidden representation $h$ using a bottleneck of size $r$, followed by a nonlinear activation function $f_{NonL}$, and an up-projection matrix $W_{up}\in \mathbb{R}^{r\times d}$ that restores the original dimension. As shown in Fig.\ref{fig:adapter}, these modules are integrated with residual connections and can operate on the output $h_{out}$ in a \textit{sequential} (Seq) manner, as follows:
\begin{equation}
\label{seq}
h_{out} = h_{out} + f_{NonL}(h_{out}W_{down})W_{up}
\end{equation}
Alternatively, they can act on the input $h$ in a \textit{parallel} (Par) manner:
\begin{equation}
\label{par}
h_{out} = h_{out} + f_{NonL}(h_{msa}W_{down})W_{up} + h_{msa}
\end{equation}
As for classification layers $W$,  we further decompose the classifier into $W = \left [ \mathbf{w}_{1}, \mathbf{w}_{2}, \cdots,\mathbf{w}_{j} \right ]$ ,where $ \mathbf{w}_{j}$ represents the classifier weight for class $j$. 

\section{Mixture of Task-specific Expert}
\label{sec:method}

\begin{figure}[t]
\centering
\includegraphics[width=1\columnwidth]{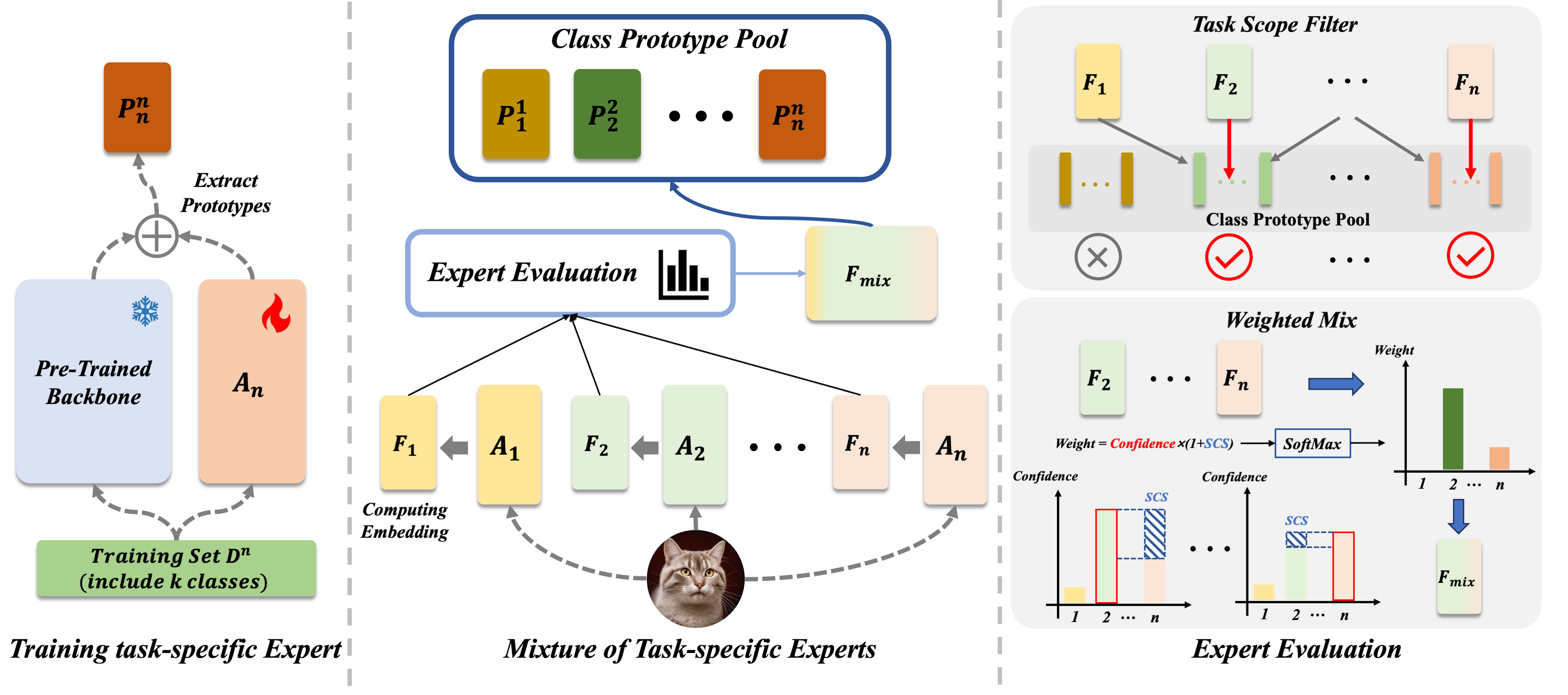}
\caption{Illustration of MoTE. \textbf{Left:} For the 
n-th task, we freeze the pre-trained backbone and train $A_n$ as the task-specific expert, extracting class prototypes. \textbf{Middle:} During inference, features extracted by different experts are compared with the class prototype pool. Expert scores are obtained using the expert evaluation strategy, and multi-expert joint inference is achieved through weighted feature fusion. \textbf{Right:} Details of the expert evaluation process. First, expert filtering is performed to retain only those experts whose predictions fall within their designated task scope. If multiple experts are selected, their scores determine their respective weights, with the highest-scoring expert receiving a greater weight. $P_{a}^{b}$ represents the class prototype generated for dataset $a$ by the $b-th$ expert; $A_{n}$ represents $n-th$ adapter; $F_{n}$ represents the features generated by the $ n-th$ adapter.}
\label{fig:overview}
\end{figure}

In this section, we first introduce the concept of Task-specific Experts, where a lightweight adapter is inserted for each sequential task. The goal is to enable the PTM to adapt to new tasks more effectively. Since training is conducted specifically for each new task, cross-task conflicts are minimized. This approach ensures that the representations of old tasks remain intact while allowing new tasks to be fully trained. Each adapter is optimized independently to become an "expert" for its corresponding task. Building upon the work of~\cite{prototype}, after training a new task, we discard the classifier layer and extract the features of all samples for each class, averaging them to generate class prototypes. To emulate the sparse expert activation and weighted feature fusion mechanisms in a Mixture of Experts (MoE), we introduce a simple yet effective expert evaluation strategy, which consists of two key components: expert filtering and the weighted mixture of expert inference. For task-ambiguous samples, the weighted mixture of the expert inference mechanism computes expert scores to weight multiple expert features, enabling high-dimensional feature fusion. An overview of our algorithm is presented in Fig.~\ref{fig:overview}. Furthermore, to explore the trade-off between the number of adapters and model performance, we extend our approach to a variant termed Adapter-limited MoTE.

\subsection{Task-specific Experts}
Pre-trained models possess strong representational capabilities, allowing them to easily adapt to downstream tasks. However, PTMs have a large number of parameters, making full fine-tuning computationally expensive and potentially damaging to the original representations. Consequently, PTM-based CIL methods typically freeze the PTM and utilize PEFT methods to learn sequential tasks. In contrast to prompt-based methods, our approach embeds lightweight adapters into the FFN layers, training each task independently. This ensures that the representations learned for previous tasks remain unaffected while new tasks are fully optimized. The parameters of the adapter are updated using the cross-entropy loss, denoted as $ \mathcal{L}_{ce} $. At task $t$, the loss is formulated as:  
\begin{equation}
\label{eq:CEloss}
\mathcal{L}_{ce, t} (\mathbf{x},y) = -\frac{1}{\left | \mathcal{D}^{t}  \right |} \sum_{(\mathbf{x} ,y)\in \mathcal{D}^{t}}  \mathbf{y}   \cdot \log p_{1:t}(\mathbf{x}) 
\end{equation}
where $\mathbf{y}$ is a one-hot encoded vector indicating the ground-truth class, with a value of 1 at the corresponding label position and 0 elsewhere. The term $p_{1:t}(\mathbf{x})$ represents the predicted probability distribution over all previously seen classes for the input $\mathbf{x}$.

The adapter acquires information about the specific characteristics of the task by optimizing the model through Eq.~\ref{eq:CEloss}, allowing it to become an expert for the given task. 

At the end of task training, the model can fit the embeddings adapted for the new task. However, since the classifiers for each task are independent and there is no access to historical data, directly merging multiple task classifiers is challenging. Following ~\cite{ADAM,EASE,prototype}, we adopt a prototype-based classifier for classification prediction. Specifically, after the completion of task training, we use the trained model to extract the embeddings for all samples of each class within the task and compute the average embedding as the class prototype:
\begin{equation}
\label{eq:prototype}
P_{t}^{j} = \frac{1}{\left | \mathcal{D}^{j}   \right | }\sum_{\mathbf{x}\in \mathcal{D}^{j} }^{} f(\mathbf{x};A_{t}(\mathbf{x})) 
\end{equation}
Here, $P_{t}^{j}$ denotes the prototype of the class $j$ in the task $t$, $\mathcal{D}^{j}$ represents all samples belonging to class $j$, and $A_{t}$ refers to the Adapter for the task $t$. Using Eq.~\ref{eq:prototype}, we obtain the class prototypes, which are then used as weights in the classifier. During inference, the model output features are compared with the prototype-based weights in the classifier using a similarity computation:
\begin{equation}
\label{eq:cosine}
\cos(v_{1},v_{2})=\frac{v_{1} \cdot v_{2}}{\left \| v_{1} \right \| \left \| v_{2} \right \| } 
\end{equation}
The cosine similarity-based classifier not only addresses the issue of classifier bias toward new classes caused by the lack of access to historical data but also facilitates the effective merging of multiple classifiers.

\subsection{Expert Filtering}
Since CIL assumes no class overlap across tasks, each sample is associated with only one task ~\cite{CILsurvey,Continualsurvey}. Ideally, only the corresponding task-specific expert should be responsible for inference and prediction. Therefore, we introduce an expert filtering strategy that discards embeddings from irrelevant experts and retains only the most relevant expert representations. Specifically, we allow all experts to participate in the forward pass and compare their outputs against all class prototypes. For expert $A_{k}$, let $C_{k}$ denote the set of class labels associated with its task. Suppose $c_{k}=\mathrm{argmax}\ \mathrm{cos}(F_{k},P^{j}_{k}) $ is the predicted class by $A_{k}$. If $c_{k} \in C_{k}$, expert $A_{k}$ is marked as a "reliable" expert; otherwise, if $c_{k} \notin C_{k}$, it is marked as "unreliable." After this filtering step, only the representations produced by reliable experts are retained for further processing. Task-aware expert filtering is illustrated on the upper right of Fig.~\ref{fig:overview}. When multiple reliable experts are identified for a given sample, it indicates that the sample is task-ambiguous. In such cases, we apply the Weighted Multi-Expert Inference mechanism to enhance classification reliability.

\subsection{Weighted Multi-Expert Inference}
There is typically a single "reliable" expert for most inference samples. However, for a small subset of ambiguous cross-task samples, multiple experts may be considered trustworthy, leading to model uncertainty during inference. To enhance the model’s discriminative ability and robustness, we evaluate expert importance using two key indicators: \textbf{Confidence} and \textbf{Self-Confidence Score (SCS)}. Based on these indicators, we assign different weights to experts during joint inference to optimize decision-making. We define the observed logit for each expert as $Z = [z_{1},z_{2},\dots,z_{n}]$.

For SCS, we compute it as the difference between the highest and second-highest logit observed by each expert, formulated as:
\begin{equation}
\label{eq:scs}
s_{i} = \frac{z^{1^{st}}_{i} - z^{2^{nd}}_{i}}{z^{1^{st}}_{i}} 
\end{equation}
where $s_{i}$ represents the self-confidence score of the $i^{th}$ expert, and $z^{n^{th}}_{i}$ denotes the $n^{th}$ highest logit value output by the $i^{th}$ expert.
The importance of each expert is then calculated as follows:
\begin{equation}
\label{eq:weight}
w_{i} = z^{1^{st}}_{i}+\gamma s_{i}
\end{equation}
Here, $\gamma$ serves as a scaling factor. However, to enhance the generalization ability of the algorithm, we set $\gamma$ to $z^{1^{st}}_{i}$, i.e., the highest confidence score, allowing the SCS adjustment to be adaptively determined rather than relying on a globally fixed hyperparameter. In Sect.~\ref{sec:gamma}, we empirically compare the performance impact of using a fixed global scaling factor versus the proposed adaptive strategy.

Thus, the final expert-based feature fusion can be formulated as:
\begin{equation}
F_{mix} = \sum_{i=1}^{n} w_{i}F_{i}
\end{equation}
where $F_{mix}$ denotes the aggregated feature representation, and $F_{i}$ is the feature of  expert $i$. The weighted expert fusion process is illustrated in the lower right corner of Fig.~\ref{fig:overview}.

\subsection{Adapter-Limited MoTE}
\label{sec:adapter_limit}
\begin{figure}[t]
\centering
\includegraphics[width=0.8\columnwidth]{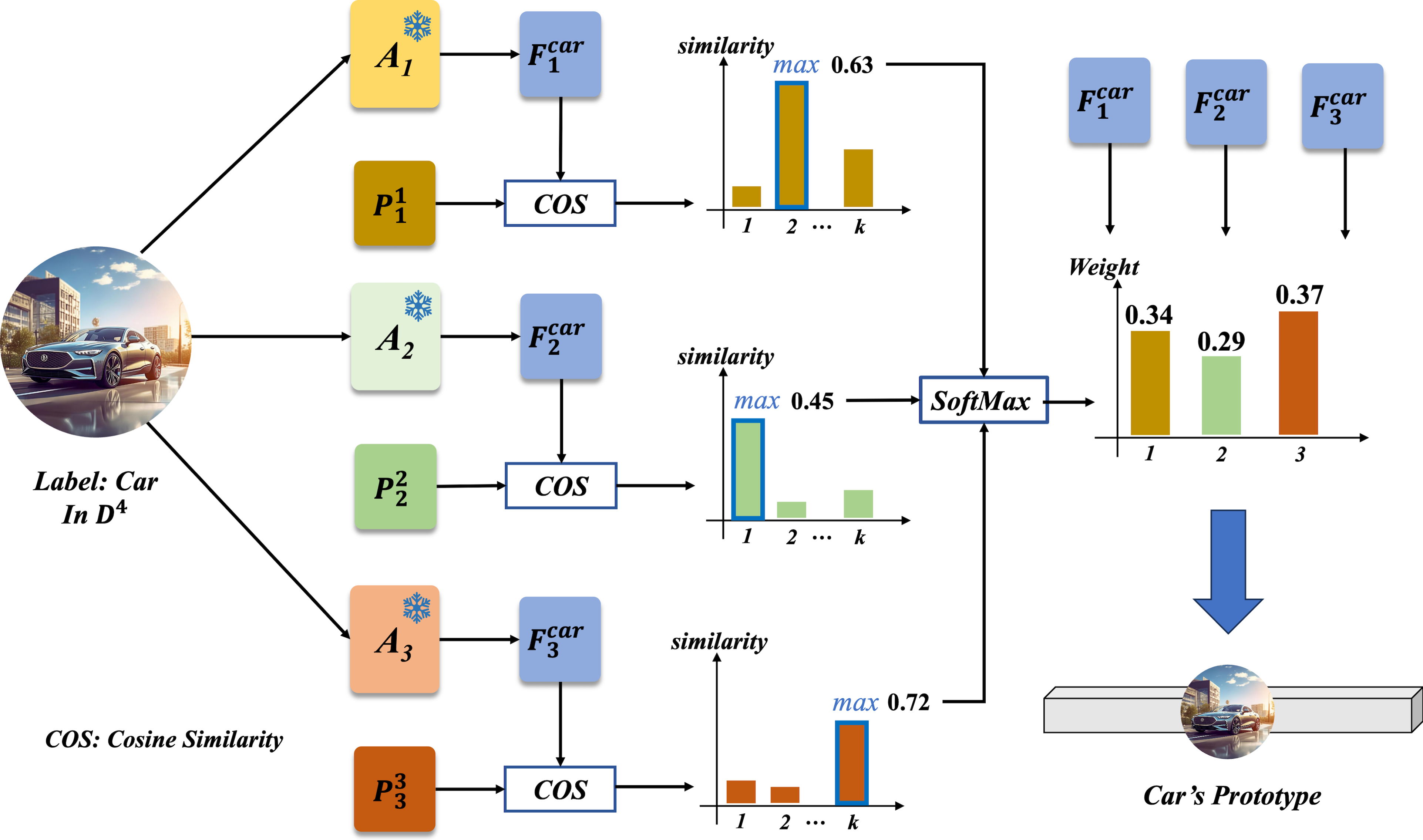}
\caption{Illustration of Adapter Limited MoTE. The number of adapters is assumed to be limited to three, with the process of generating class prototypes for the fourth task used as an example. Here, $D^{4}$ represents the dataset of the fourth task, and \textbf{\textit{COS}} denotes the similarity computation using Eq.\ref{eq:cosine}.}
\label{fig:adapter_limited}
\end{figure}

To investigate the impact of the number of adapters on model performance, we propose \textbf{Adapter-Limited MoTE} based on the original MoTE framework. Specifically, when the number of tasks is less than or equal to the adapter limit, newly added adapters are independently trained on samples from new tasks, becoming task-specific experts. When the number of tasks exceeds the limit, data from the new task is passed through the existing adapters for inference. The features of all samples in the new classes are averaged to create adapter-specific class prototypes. These prototypes are compared with all class prototypes within the task scope of each adapter using similarity metrics, retaining the highest similarity value. This process is repeated for all adapters, and the highest similarity scores are then weighted and merged to form the class prototype for the new class, as illustrated in Fig.~\ref{fig:adapter_limited}. The inference process aligns with the expert evaluation mechanism proposed in this study, where samples outside the task scope are combined using confidence-weighted aggregation.  We conducted detailed ablation studies, with the experimental results presented in Sec.~\ref{Sec:ablation_limit}.

\section{Experiment}
In this section, we first introduce the experimental setup and implementation details, along with the baseline algorithms used for comparison. We then conduct extensive experiments across five datasets and five random seeds to validate the superior performance of MoTE. Next, we perform detailed ablation studies to assess the effectiveness of different components and compare the performance of two different adapter embedding strategies. Finally, we provide in-depth analyses, including inference time comparisons, visualizations, and evaluations of Adapter-Limited MoTE across multiple datasets under different adapter constraints.

\subsection{Implementation Details}
\subsubsection{Datasets}
We followed the research of~\cite{l2p,ADAM,EASE} and selected five commonly used datasets to evaluate the algorithm’s performance: CIFAR100~\cite{CIFAR}, CUB~\cite{CUB}, ImageNet-A~\cite{ImageNetA}, ImageNet-R~\cite{ImageNetR}, and VTAB~\cite{VTAB}. Specifically, CIFAR100 consists of 100 classes, CUB, ImageNet-A, and ImageNet-R each contain 200 classes, and VTAB includes 50 classes. For clarity, in the figures and tables of this paper, ImageNet-A and ImageNet-R are abbreviated as \textbf{INA} and \textbf{INR}, respectively.

\subsubsection{Dataset Split}
Following the benchmark setup of ~\cite{icarl,CILsurvey,EASE,ADAM}, we denote the class split as "B-$m$ Inc-$n$". Here, $m$ represents the number of classes in the first stage, and $n$ indicates the number of classes in each subsequent incremental stage, and the scope of the incremental task is $n$. We followed prior research to ensure a fair comparison and set the random seed to \{1993\}. To provide a more generalized algorithm performance evaluation, we conducted experiments on five different random seeds \{1991, 1992, 1993, 1994, 1995\} across various datasets in the main performance comparison experiments.

\subsubsection{Baseline Methods}
This subsection introduces the baseline algorithms used for comparison in this paper, categorized into exemplar-free and exemplar-based methods. The exemplar-free approaches do not rely on any samples from previous tasks. Below is a list of exemplar-free algorithms:
\begin{itemize}
    \item LwF~\cite{lwf}: is the first to apply knowledge distillation to CIL. By minimizing the difference between the outputs of the new and old models, LwF preserves knowledge from previous tasks while learning from new task data.
    \item iCaRL~\cite{icarl}: employs knowledge distillation and exemplar replay to retain previously learned knowledge. Additionally, it utilizes a nearest-class-mean classifier for final classification.
    \item DER~\cite{DER}: expands the network for class-incremental learning by freezing the previous backbone and initializing a new one for each task. It concatenates features from all historical backbones and employs a large linear classifier, requiring exemplars for calibration. While achieving strong performance, DER incurs high memory overhead due to storing all past backbones.
    \item Foster~\cite{Foster}: reduces the memory overhead of DER by compressing backbones through knowledge distillation. Instead of storing multiple backbones, it maintains a single backbone throughout the learning process while achieving feature expansion with minimal memory cost.
    \item SimpleCIL~\cite{ADAM}: Explores a prototype-based classifier using a vanilla pre-trained model. It initializes with a PTM, builds a prototype classifier for each class, and employs a cosine similarity classifier for classification.
    \item Adam~\cite{ADAM}: Extends SimpleCIL by combining the pre-trained and adapted models. It treats the first incremental stage as the sole adaptation phase, adapting the PTM to extract task-specific features, thus unifying generalizability and adaptivity within a single framework. Its variants include Adam\_\{Finetune, VPT\_S, VPT\_D, SSF, Adapter\}.
    \item L2P~\cite{l2p}: The first method to incorporate pre-trained Vision Transformers into continual learning. It freezes pre-trained weights and employs visual prompt tuning~\cite{vpt} to capture new task features. Instance-specific prompts are constructed from a prompt pool using key-value mapping.
    \item DualPrompt~\cite{dualprompt}: is an extension of L2P, introducing two types of prompts: general and expert prompts. Other details remain the same as L2P, using a prompt pool to construct instance-specific prompts.
    \item CODA-Prompt~\cite{codaprompt}: addresses the limitation of prompt selection and eliminates this process through prompt reweighting, replacing selection with attention-based prompt recombination.
    \item EASE~\cite{EASE}: trains lightweight adapter modules for each task, creating task-specific subspaces for joint decision-making and uses a semantic-guided prototype complement strategy to synthesize new features for old classes without requiring old instances.
\end{itemize}

\subsubsection{Training Details}
We implement our method using PyTorch~\cite{pytorch} and conduct experiments on two NVIDIA GeForce RTX 3090 GPUs. The experiments leveraged the public implementations of existing CIL methods within the PILOT framework~\cite{Pilot}. As for PTM, we follow the work of Wang et al.~\cite{l2p,dualprompt} and Zhou et al.~\cite{ADAM,EASE}, we use representative models, namely {\bf ViT-B/16-IN21K} as the PTM, which is pre-trained on ImageNet21K. During training, we employ Stochastic Gradient Descent (SGD) as the optimizer with an initial learning rate of 0.01 and a weight decay of 0.005 to mitigate overfitting. The model is trained for 20 epochs with a batch size of 48 to ensure stable gradient updates. To enhance convergence and prevent premature stagnation, we utilize a \textbf{cosine annealing} learning rate scheduler, which progressively decays the learning rate to a minimum value of 0 for training.

\begin{figure}[!t]
\centering
	\begin{subfigure}{0.32\linewidth}
		\includegraphics[width=1\columnwidth]{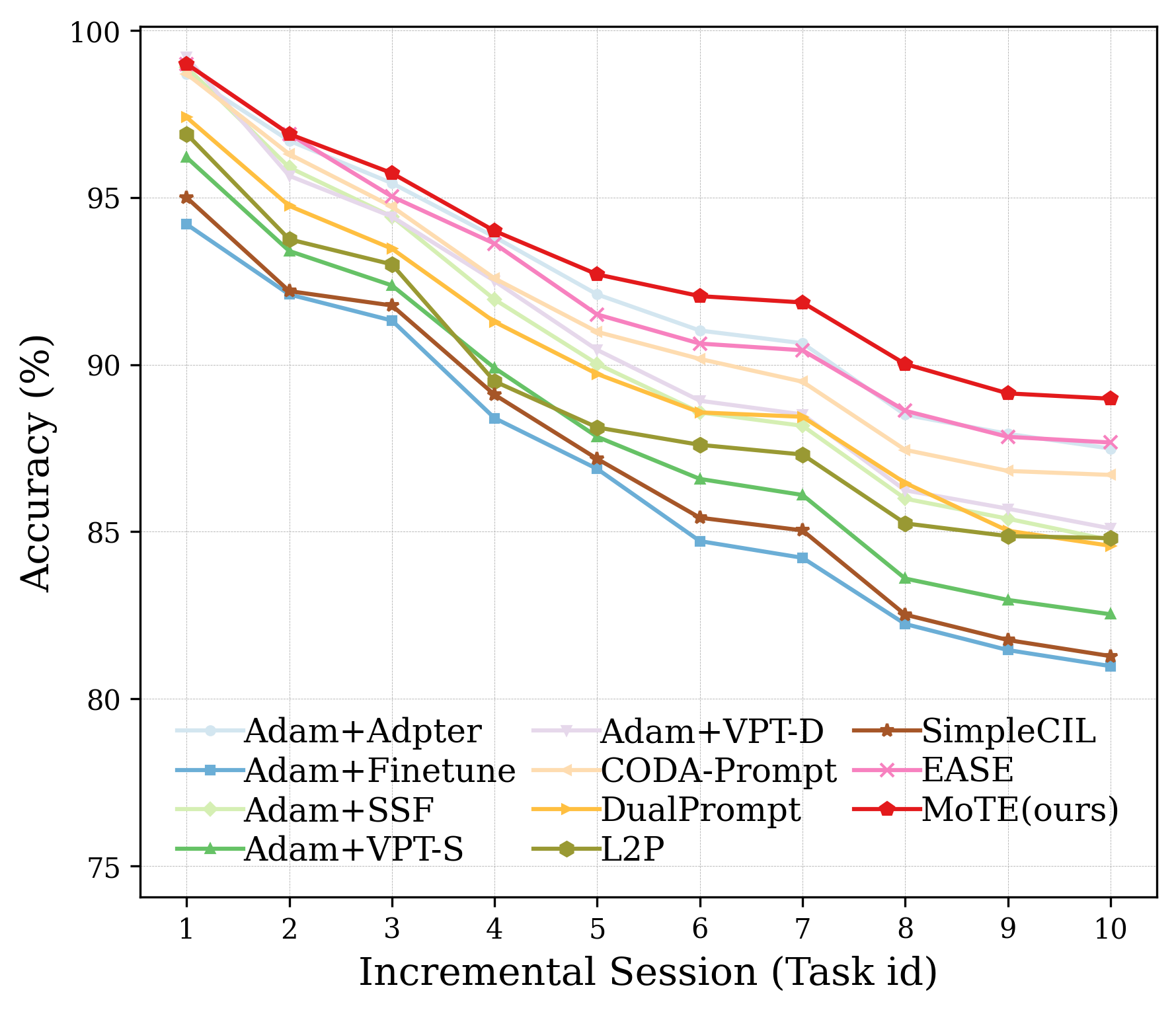}
            \vspace{-7mm}
		\caption{CIFAR B0-Inc10}
		\label{fig:benchmark-cifar-10-in21k}
	\end{subfigure}
	\hfill
	\begin{subfigure}{0.32\linewidth}
		\includegraphics[width=1\linewidth]{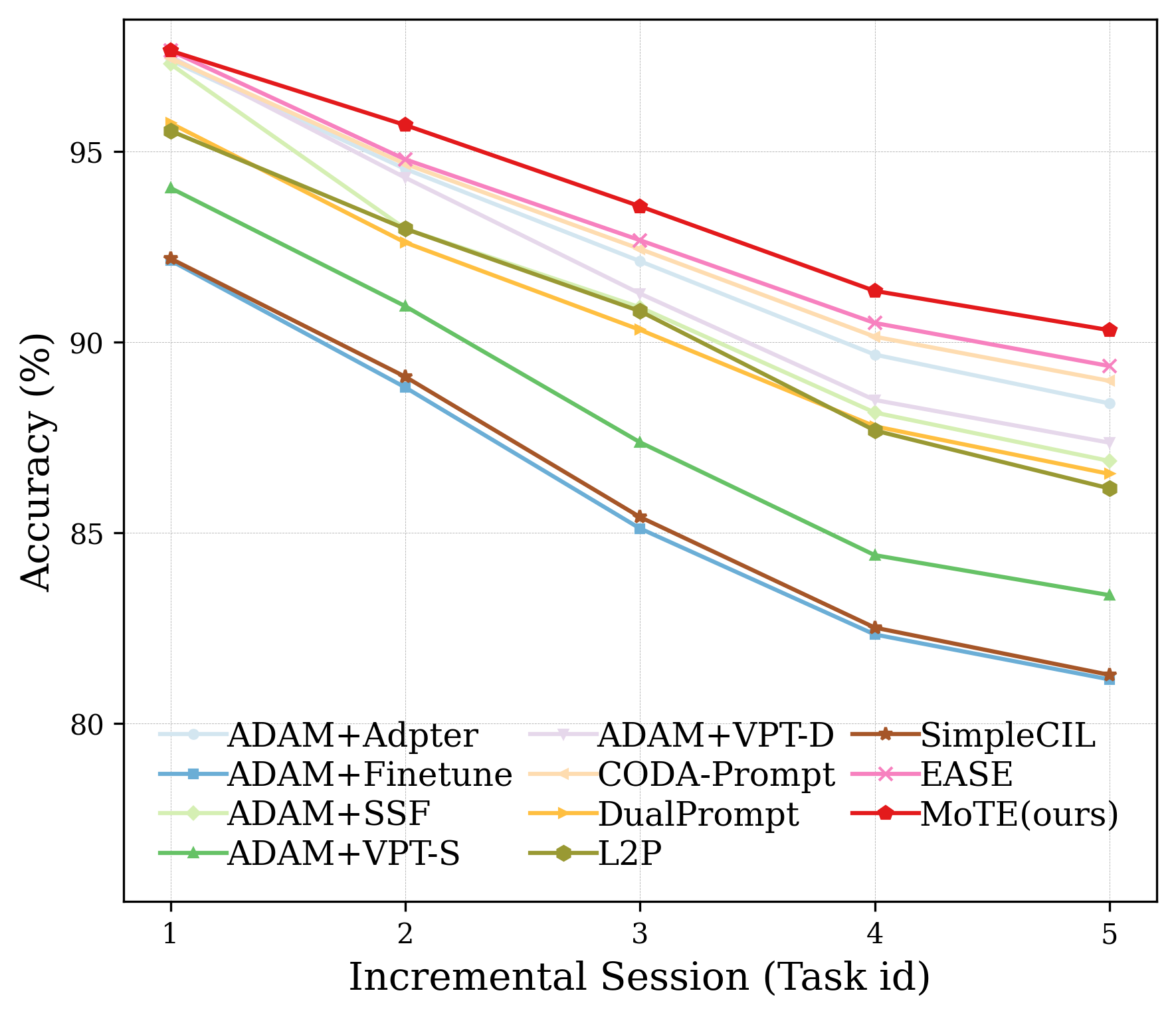}
            \vspace{-7mm}
		\caption{CIFAR B0-Inc20}
		\label{fig:benchmark-cifar-20-in21k}
	\end{subfigure}
        \hfill
        \begin{subfigure}{0.32\linewidth}
		\includegraphics[width=1\linewidth]{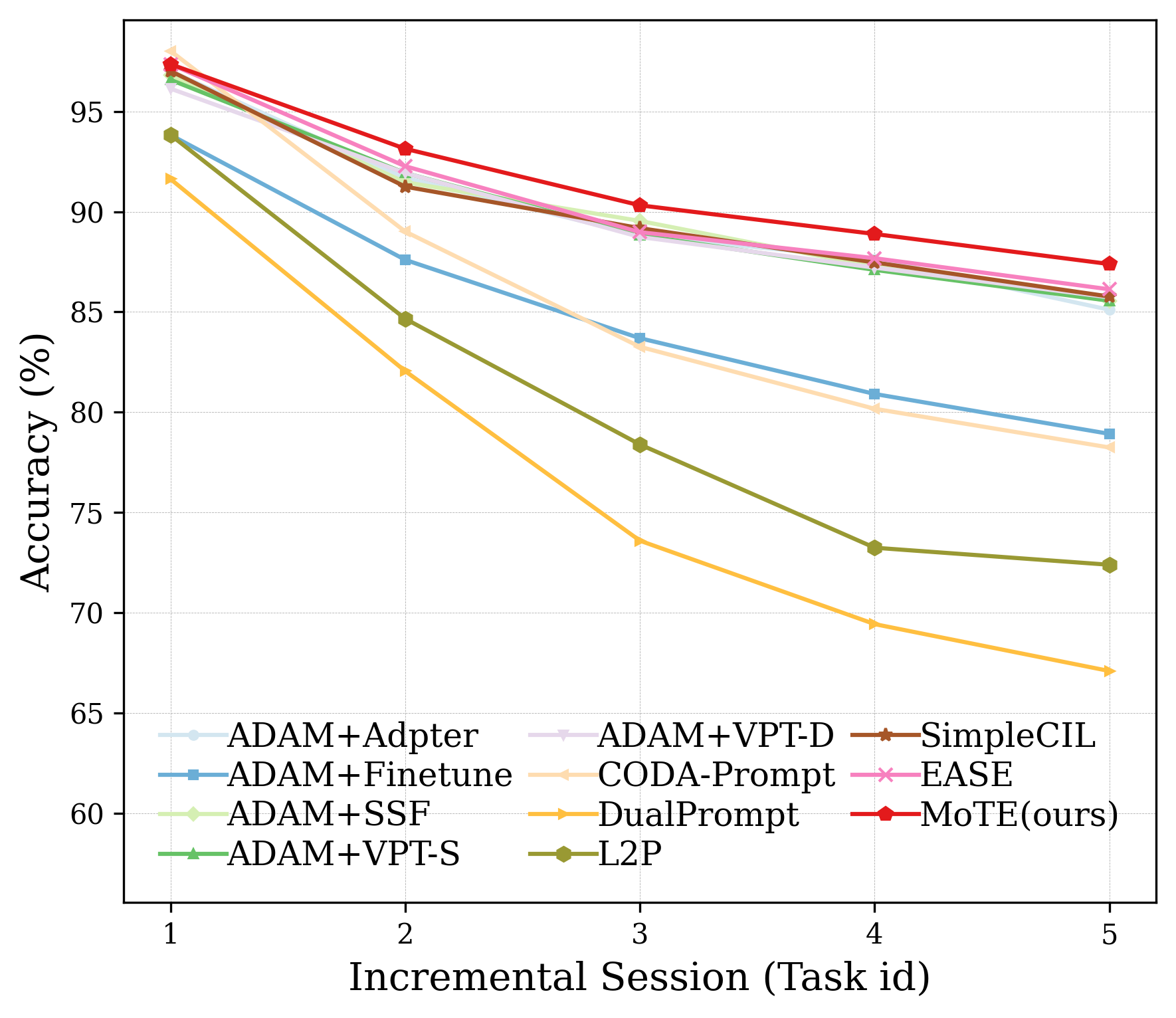}
            \vspace{-7mm}
		\caption{CUB B0-Inc40}
		\label{fig:benchmark-cub-40-in21k}
	\end{subfigure}
        
        \begin{subfigure}{0.32\linewidth}
		\includegraphics[width=1\linewidth]{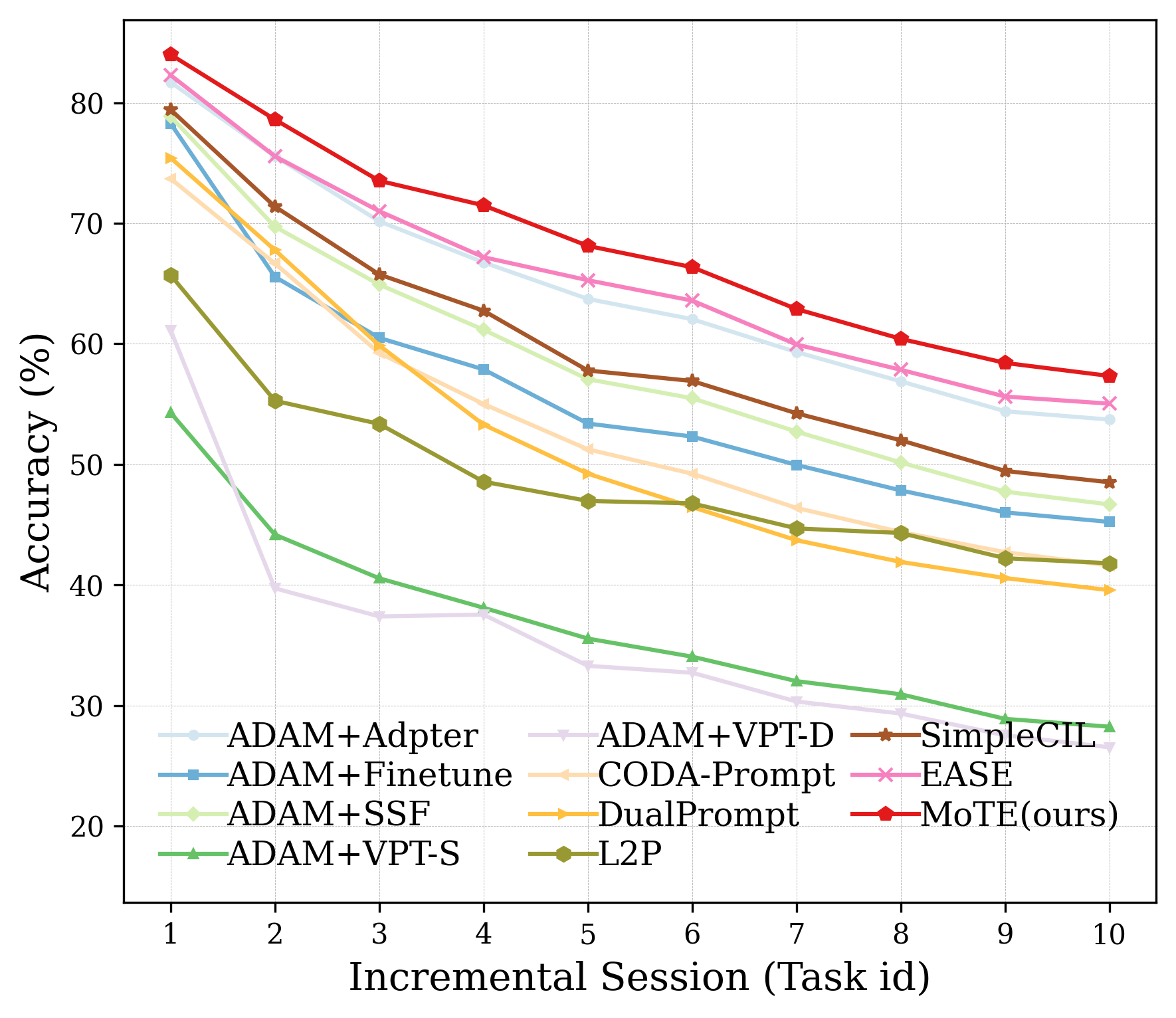}
            \vspace{-7mm}
		\caption{INA B0-Inc20}
		\label{fig:benchmark-ina-20-in21k}
	\end{subfigure}
        \hfill
        \begin{subfigure}{0.32\linewidth}
		\includegraphics[width=1\linewidth]{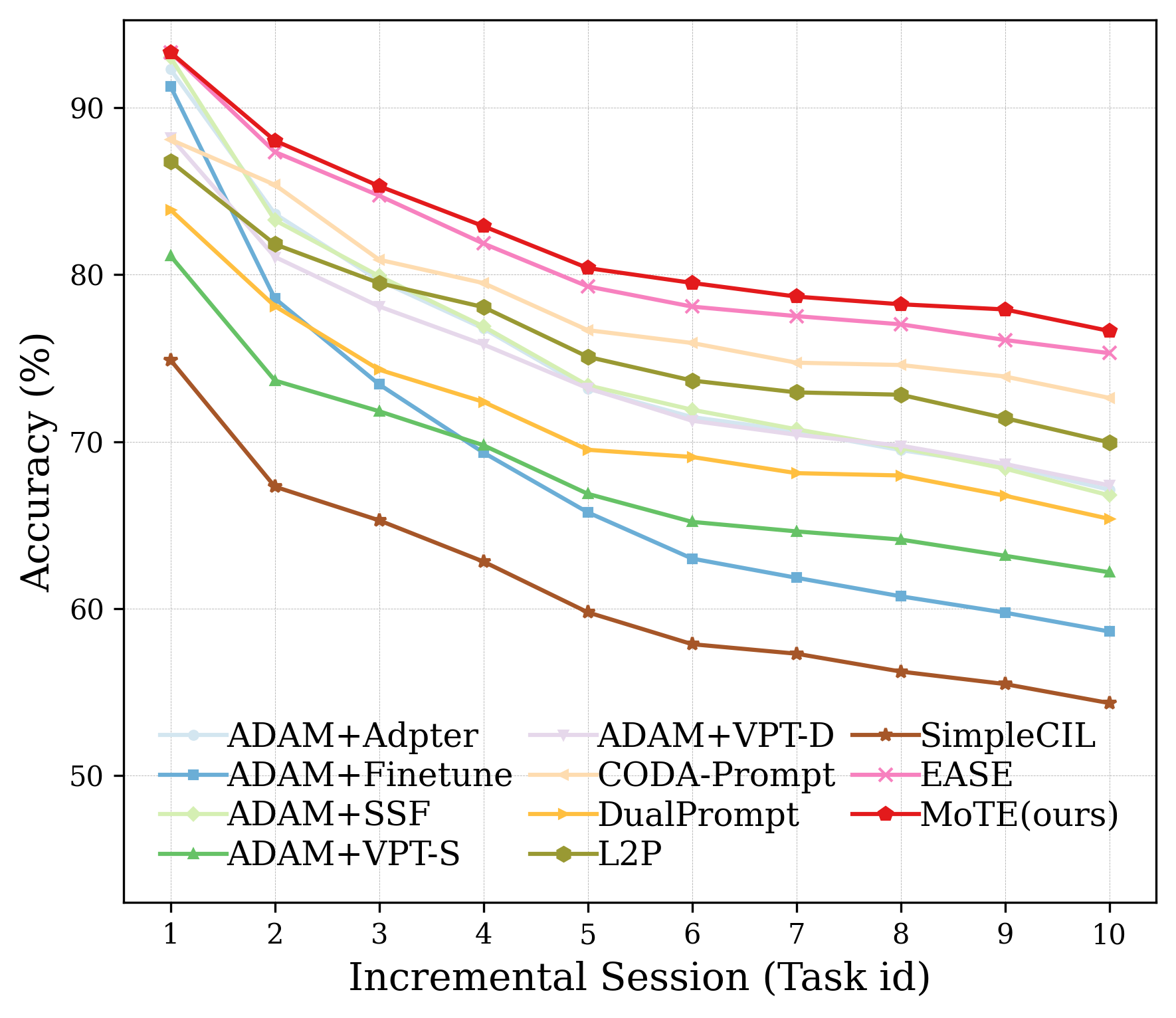}
            \vspace{-7mm}
		\caption{INR B0-Inc20}
		\label{fig:benchmark-inr-20-in21k}
	\end{subfigure}
        \hfill
        \begin{subfigure}{0.32\linewidth}
		\includegraphics[width=1\linewidth]{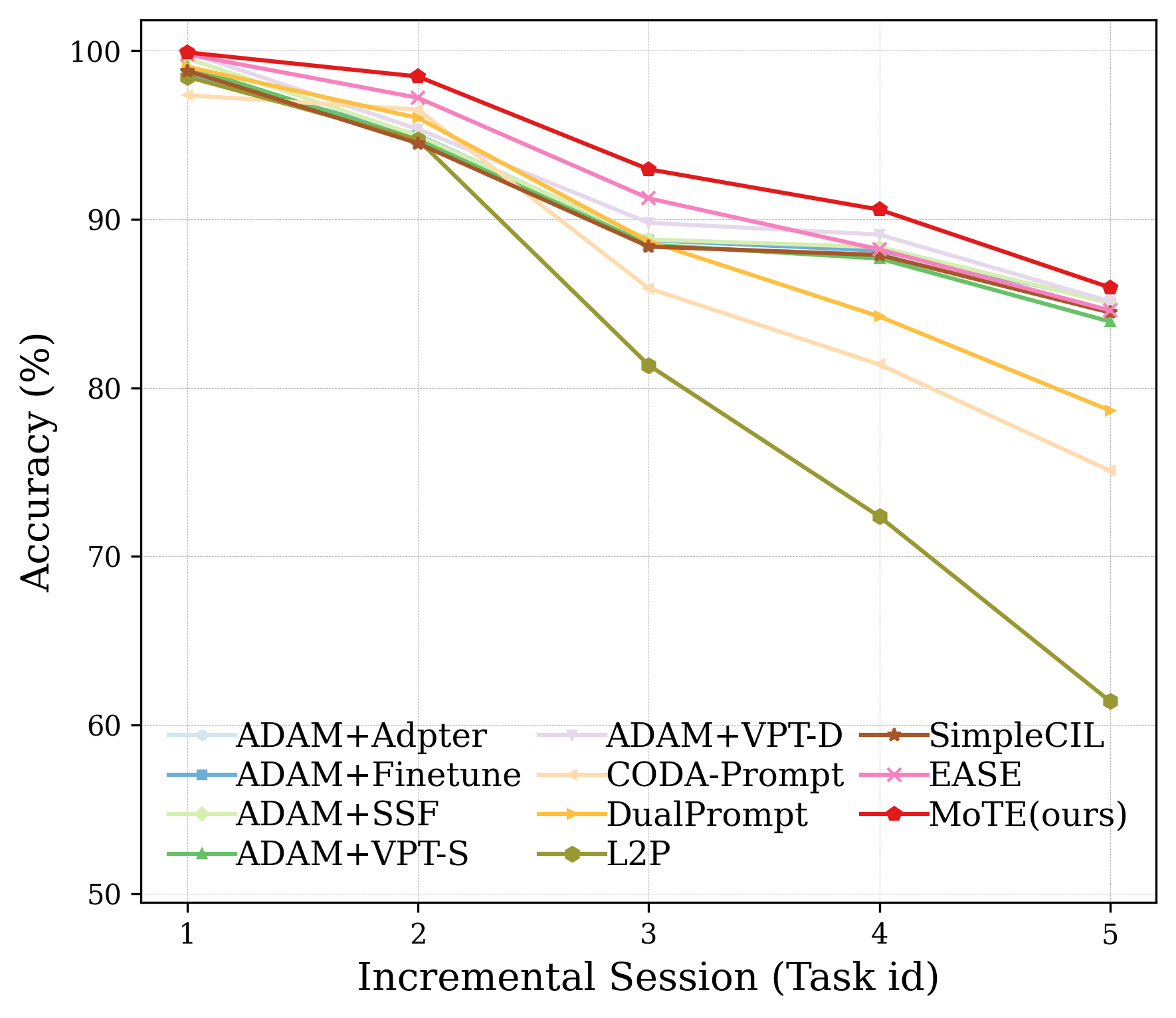}
            \vspace{-7mm}
		\caption{VTAB B0-Inc10}
		\label{fig:benchmark-vtab-10-in21k}
	\end{subfigure}

        \begin{subfigure}{0.32\linewidth}
		\includegraphics[width=1\linewidth]{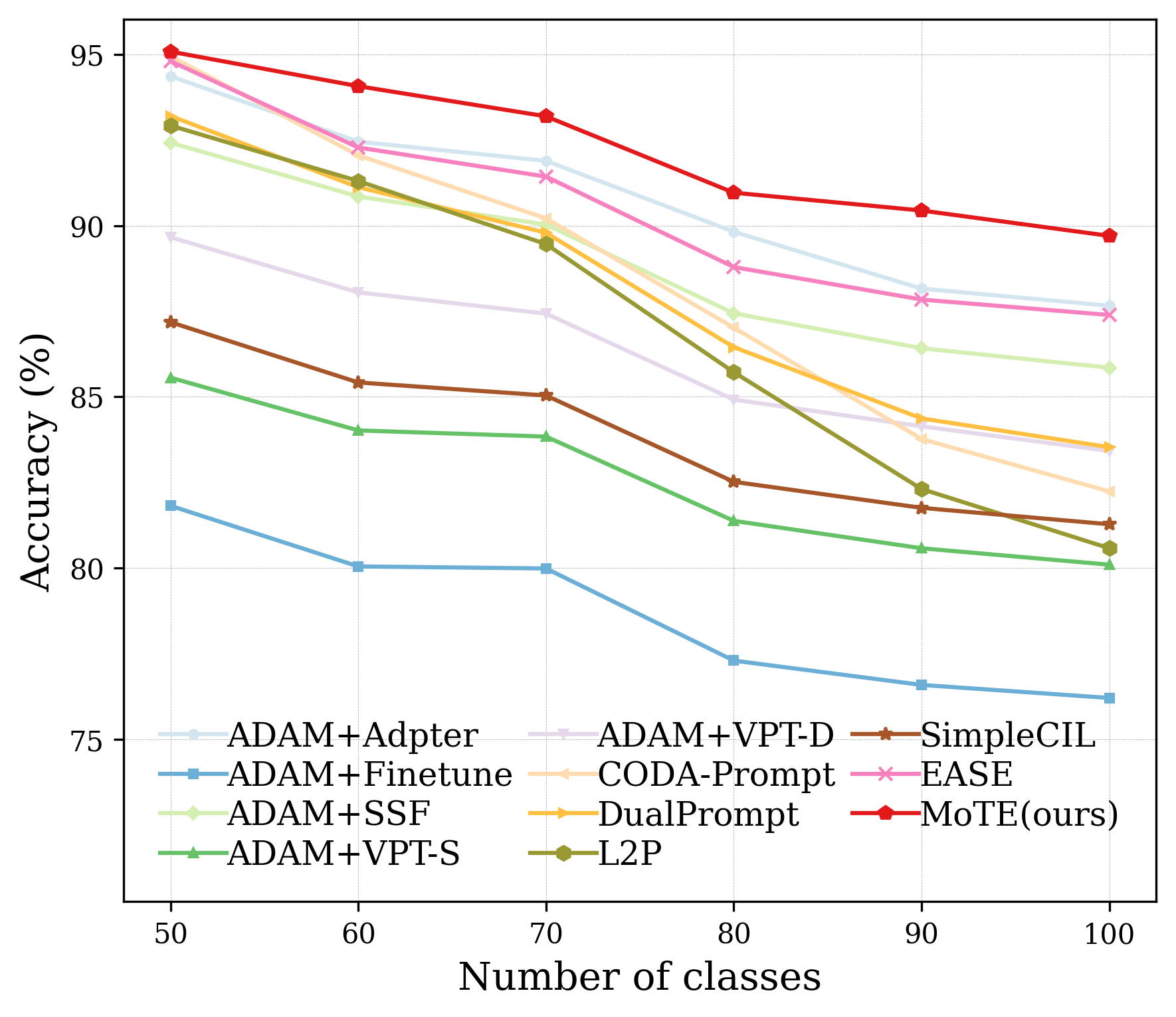}
            \vspace{-7mm}
		\caption{CIFAR B50-Inc10}
		\label{fig:benchmark-cifar-50-in21k}
	\end{subfigure}
        \hfill
        \begin{subfigure}{0.32\linewidth}
		\includegraphics[width=1\linewidth]{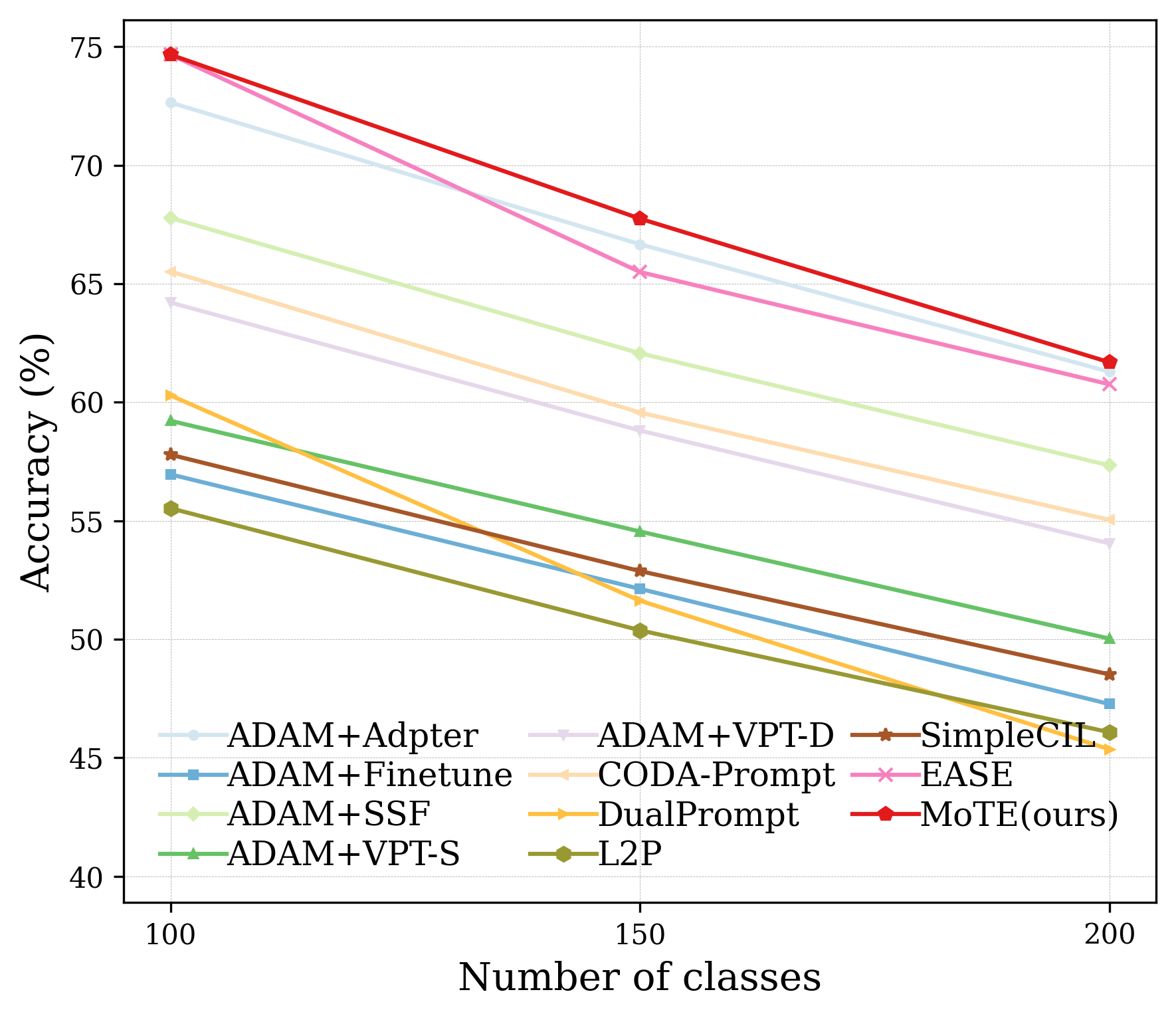}
            \vspace{-7mm}
		\caption{INA B100-Inc50}
		\label{fig:benchmark-ina-100-in21k}
	\end{subfigure}
        \hfill
        \begin{subfigure}{0.32\linewidth}
		\includegraphics[width=1\linewidth]{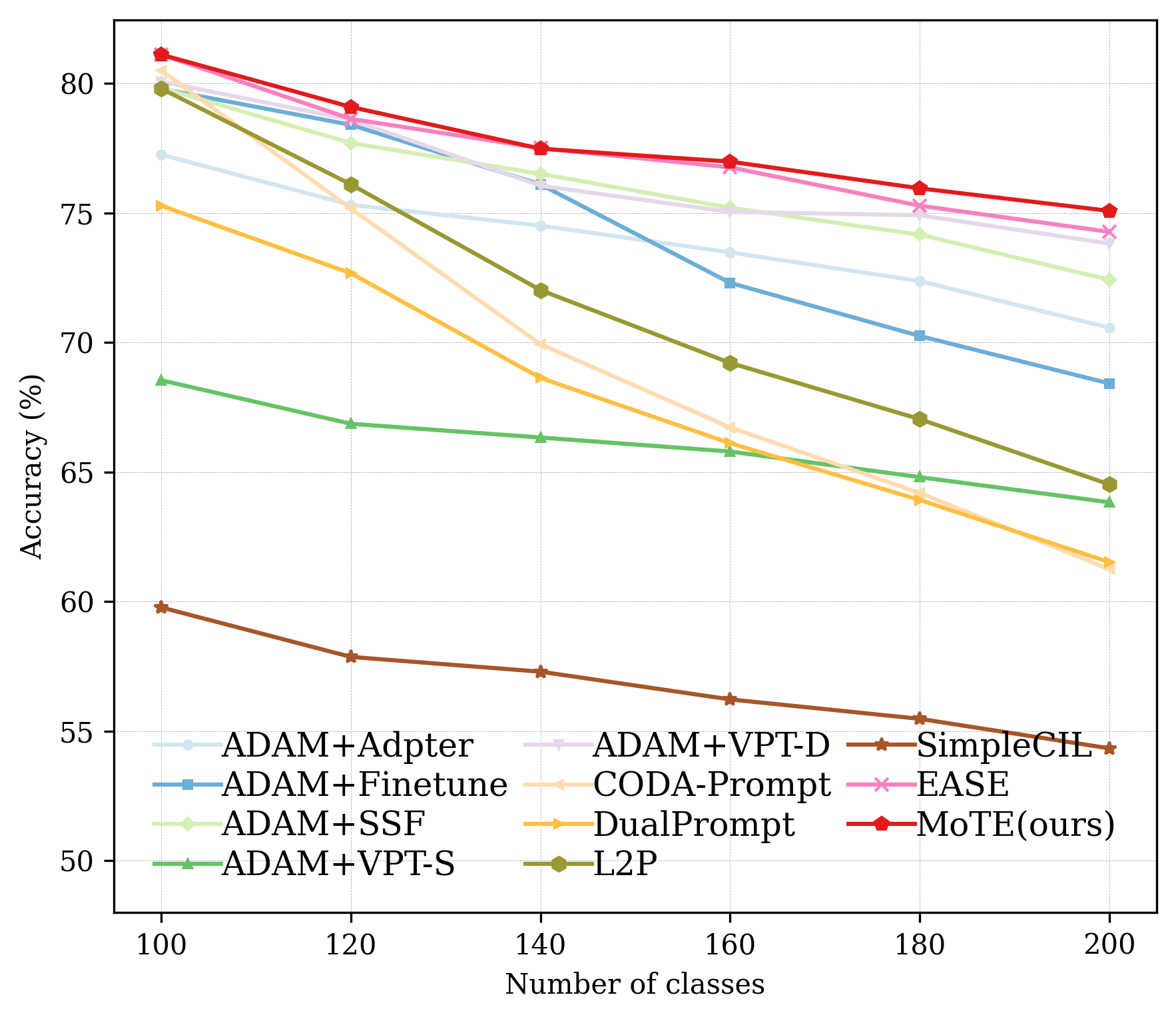}
            \vspace{-7mm}
		\caption{INR B100-Inc20}
		\label{fig:benchmark-inr-100-in21k}
	\end{subfigure}
    
        \vspace{-2mm}
	\caption{\small Performance comparison of different methods across various experimental settings, all initialized with ({\bf ViT-B/16-IN21K}).}
	\label{fig:benchmark-in21k}
\end{figure}

\subsubsection{Evaluation metric}
In CIL methods, we use standard metrics to evaluate performance: "Last" refers to the accuracy on the most recent task, and "Avg" refers to the Average Accuracy, which calculates the accuracy across all observed classes~\cite{Continualsurvey}. Let $a_{i,j}$ be the accuracy of the model on the test set of task $j$ after training from task 1 to task $i$. The Average Accuracy can be calculated as follows:
\begin{equation}
\label{avg}
Avg(A_{T}) = \frac{1}{T} \sum_{j=1}^{T} a_{T,j}   
\end{equation}

The extent of forgetting in the model can be evaluated using the Average Forgetting (AF) metric. Specifically, the forgetting for a task is defined as the difference between its peak performance achieved in the past and its performance at the current evaluation stage:
\begin{equation}
\label{af}
f_{j,k}=\max_{i\in \{1,...,k-1 \}}  (a_{i,j}-a_{k,j} ),\forall j< k  
\end{equation}
The measurement of Average Forgetting (AF) across all previous tasks can be expressed as:
\begin{equation}
AF_{T} = \frac{1}{T-1}\sum_{j=1}^{T-1}f_{T,j} 
\end{equation}

\subsection{Benchmark Comparison}
\label{Sec:benchmark}

\begin{table}[t]
\caption{Comparison of different methods under various experimental settings in terms of Average Forgetting (AF) and Average Accuracy (Avg). All algorithms use \textbf{ViT-B/16-IN21K} as the pre-trained model. Bold text indicates superior performance.}
\label{tab:var}
\centering
\vspace{-2mm}
\resizebox{\textwidth}{!}{
\begin{tabular}{lcccccccccc}
\hline
\multirow{2}{*}{Method} &
  \multicolumn{2}{c}{CIFAR100 B0-Inc10} &
  \multicolumn{2}{c}{CUB B0-Inc20} &
  \multicolumn{2}{c}{INA B0-Inc20} &
  \multicolumn{2}{c}{INR B0-Inc20} &
  \multicolumn{2}{c}{VTAB B0-Inc10} \\ \cline{2-11} 
               & Avg($\uparrow$)        & AF($\downarrow$)        & Avg($\uparrow$)        & AF($\downarrow$)          & Avg($\uparrow$)        & AF($\downarrow$)          & Avg($\uparrow$)        & AF($\downarrow$)         & Avg($\uparrow$)        & AF($\downarrow$)          \\ \hline
L2P            & 88.50±0.64 & 9.06±1.62 & 77.91±1.31 & 11.10±3.83 & 49.09±1.70 & \textbf{6.88}±2.10  & 76.35±0.35 & 4.96±0.84 & 77.56±4.68 & 25.45±9.33 \\
Dualprompt     & 88.74±0.48 & 4.35±0.71 & 83.87±0.56 & 7.96±1.87  & 52.85±1.11 & 9.86±0.91  & 71.56±0.38 & \textbf{3.89}±0.60 & 86.23±3.45 & 12.98±4.62 \\
CODA\_Prompt   & 91.44±0.12 & 4.72±0.56 & 83.45±1.33 & 5.85±1.52  & 52.18±1.63 & 8.35±1.73  & 77.87±0.49 & 3.95±0.78 & 85.87±3.72 & 13.07±5.06 \\
SimpleCIL      & 86.18±0.56 & 5.69±0.50 & 90.75±1.02 & 5.04±0.65  & 58.99±0.66 & 11.71±0.80 & 61.45±0.75 & 7.94±0.85 & 90.69±0.64 & 7.52±2.54  \\
Adam\_Finetune & 73.09±4.88 & 6.47±0.79 & 88.70±0.93 & 5.66±0.43  & 54.55±2.71 & 11.70±0.97 & 68.21±0.49 & 5.94±0.77 & 90.50±1.04 & \textbf{7.22}±2.14  \\
Adam\_VPT\_S   & 82.84±4.66 & 6.52±1.31 & 90.07±1.13 & 5.35±0.86  & 58.47±1.85 & 11.54±0.77 & 68.62±0.54 & 6.66±0.47 & 90.01±1.45 & 7.64±2.17  \\
Adam\_VPT\_D   & 86.71±1.00 & 6.02±0.40 & 90.07±1.13 & 5.35±0.86  & 54.04±3.31 & 10.97±1.20 & 73.75±0.68 & 6.77±0.32 & 90.83±0.95 & 7.57±1.54  \\
Adam\_SSF      & 90.29±0.76 & 4.51±0.38 & 90.57±0.78 & 5.31±0.85  & 54.74±4.58 & 10.97±1.20 & 74.89±0.65 & 6.82±0.55 & 91.05±1.48 & 7.48±2.43  \\
Adam\_Adapter  & 91.57±0.66 & \textbf{4.35}±0.23 & 90.96±0.84 & \textbf{4.83}±0.63  & 59.27±0.67 & 11.51±0.76 & 75.34±0.60 & 6.53±0.52 & 90.67±0.65 & 7.56±2.49  \\
EASE           & 92.40±0.30 & 6.33±0.96 & 89.91±0.77 & 7.20±0.55  & 64.36±1.26 & 14.05±2.29 & 81.42±0.58 & 7.70±0.63 & 90.32±2.95 & 9.66±4.10  \\ \hline
MoTE           & \textbf{92.95}±0.22 & 4.70±0.52 & \textbf{91.74}±0.69 & 5.06±0.66  & \textbf{67.26}±0.96 & 11.19±1.69 & \textbf{81.93}±0.53 & 6.91±0.41 & \textbf{91.43}±1.77 & 7.54±2.44  \\ \hline
\end{tabular}
    }
\end{table}

In this section, we compare the performance of MoTE with other state-of-the-art methods. To evaluate the robustness of the approach, we used five datasets with commonly adopted protocols and conducted experiments with five random seeds. The average and standard deviation of the Avg and AF metrics over five runs were computed. To ensure a fair comparison, all algorithms were based on the same pre-trained model (ViT-B/16-IN21K). 

As shown in Tab.\ref{tab:var}, MoTE performs best under all experimental settings. The bolded results indicate the optimal performance for each setting, demonstrating the effectiveness and robustness of our method. Although MoTE achieves the best average accuracy, its performance on the AF metric does not appear to be state-of-the-art. This can be attributed to the zero-shot capability of the PTM, which ensures a lower-bound performance for the entire model, even when the algorithm itself underperforms. As a result, some algorithms, such as L2P and Dualprompt, may exhibit suboptimal test performance while maintaining relatively low performance fluctuations, leading to lower AF values. Compared to algorithms with similar performance, such as EASE, MoTE not only achieves superior average accuracy but also demonstrates improved results on the AF metric, highlighting its robustness and effectiveness. We report the incremental performance trends of various methods in Fig.~\ref{fig:benchmark-in21k}, where our algorithm demonstrates competitive performance on each incremental task. We also set up a large-scale base task experiment, as shown in Fig.~\ref{fig:benchmark-cifar-50-in21k}, Fig.~\ref{fig:benchmark-ina-100-in21k}, and Fig.~\ref{fig:benchmark-inr-100-in21k}, where MoTE consistently outperforms other algorithms at each stage. Furthermore, to demonstrate that our method surpasses the generalization capability of the PTM, we select three representative CIL approaches as baselines and conduct experiments using the same PTM. The results presented in Tab.\ref{tab:exemplars} indicate that MoTE consistently outperforms these baselines, further validating its effectiveness.

\begin{table}[t]
\caption{Comparison to traditional CIL methods. MoTE does not use any exemplars. All methods are based on the same pre-trained model (ViT-B/16-IN21K).}
\label{tab:exemplars}
\vspace{-2mm}
\centering
% \resizebox{\textwidth}{!}{
\begin{tabular}{lccccccc}
\hline
       &           & \multicolumn{2}{c}{CIFAR B0-Inc10} & \multicolumn{2}{c}{CUB B0-Inc20} & \multicolumn{2}{c}{INR B0-Inc20} \\ \cline{3-8} 
Method & Exemplars & Avg              & Last            & Avg             & Last           & Avg             & Last           \\ \hline
LwF & 0/class &86.73 &78.33 &79.98 &66.75 & 78.55 & 68.62 \\
iCaRL  & 20/class  & 84.92            & 75.17           & 87.49           & 81.21          & 71.84           & 61.72          \\
DER    & 20/class  & 88.63            & 79.18           & 84.76           & 79.09          & 80.89           & 74.82          \\
FOSTER & 20/class  & 91.91            & 87.55           & 80.85           & 78.20          & 81.37           & 75.15          \\ \hline
MoTE   & 0/class         & \textbf{93.07}   & \textbf{88.96}  & \textbf{91.83}  & \textbf{86.77} & \textbf{82.07}  & \textbf{76.28} \\ \hline
\end{tabular}
% }
\end{table}

\subsection{Ablation Study}
\label{Sec:ablation}
\subsubsection{Ablation Study on MoTE Components}
\begin{table}[t]
\caption{Ablation study on the components of MoTE. All experiments are conducted using the same pre-trained model \textbf{(ViT-B/16-IN21K)}. Confidence Re-weighting and Self-Confidence Score Re-weighting are abbreviated as \textbf{\textit{C-R}} and \textbf{\textit{SCS-R}},
respectively. Bold text indicates superior performance.}
\label{tab:abla}
\vspace{-2mm}
\centering
\resizebox{\textwidth}{!}{
\begin{tabular}{lccccccccccccccc}
\hline
\multirow{2}{*}{\#} &
\multirow{2}{*}{Filtering} &
  \multirow{2}{*}{\textit{C-R}} &
  \multirow{2}{*}{\textit{SCS-R}} &
  \multicolumn{2}{c}{CIFAR100 B0-Inc10} &
  \multicolumn{2}{c}{CUB B0-Inc20} &
  \multicolumn{2}{c}{INA B0-Inc20} &
  \multicolumn{2}{c}{INR B0-Inc20} &
  \multicolumn{2}{c}{VTAB B0-Inc10} \\ \cline{5-14} 
  & &   &   & Last  & Avg   & Last  & Avg   & Last  & Avg   & Last  & Avg & Last  & Avg   \\ \hline
1 &$\times$ &$\times$  &$\times$  &87.69 &92.46 &82.45 &88.33 &53.51 &65.98 &73.47 &80.48 &76.15 &88.99\\
2& \checkmark & $\times$  &  $\times$ & 88.18 & 92.82 & 84.88 & 90.87 & 56.09 & 67.16 & 74.15 & 81.07  & 81.09 & 91.64 \\
 3& \checkmark& \checkmark & $\times$ & 88.63 & 92.87 & 85.79 & 91.30 & 56.34 & 67.90 & 76.12 & 81.93  & 84.71 & 92.78 \\
  4&  \checkmark& $\times$& \checkmark  & 88.45 & 92.90 & 85.28 & 91.23 & 56.55 & 67.37 & 74.97 & 81.35  & 82.94 & 92.35 \\
5 & \checkmark &
  \checkmark &
  \checkmark &
  \textbf{88.92} &
  \textbf{93.06} &
  \textbf{86.77} &
  \textbf{91.90} &
  \textbf{57.93} &
  \textbf{68.44} &
  \textbf{76.28} &
  \textbf{82.07} &
  \textbf{85.22} &
  \textbf{93.16} \\ \hline
\end{tabular}
}
\end{table}

We conducted ablation studies on the three key components of MoTE. For clarity, we denote different configurations as follows: \textbf{(\#1)} taking the direct average of all experts’ inference features; \textbf{(\#2)} task filtering by experts, which eliminates experts whose predictions fall outside the task's scope; \textbf{(\#3)} weighted mixing of all experts based on their confidence levels; \textbf{(\#4)} using only SCS for weighting without considering confidence; and \textbf{(\#5)} the complete MoTE model. The multi-experiment results in Tab.~\ref{tab:abla} validate the effectiveness of each component. In the INR B0-Inc20 setting, we observe a slight performance drop after incorporating the SCS component. This is likely due to incorrect experts exhibiting excessive confidence, causing a bias in the joint inference weights when handling task-confused samples. In all other experimental settings, the best performance is achieved when all three components are utilized together.

\subsubsection{Comparison of Task-identify Accuracy}

\begin{table}[t]
\caption{Performance comparison of task identify accuracy. Task-identify Accuracy is abbreviated as \textbf{TIA}. Bold text indicates superior performance.}
\label{tab:TaskidCompare}
\centering
\vspace{-2mm}
\resizebox{\textwidth}{!}{
\begin{tabular}{lcccccccccc}
\hline
\multirow{2}{*}{Method} &
  \multicolumn{2}{c}{CIFAR100 B0-Inc10} &
  \multicolumn{2}{c}{CUB B0-Inc20} &
  \multicolumn{2}{c}{INA B0-Inc20} &
  \multicolumn{2}{c}{INR B0-Inc20} &
  \multicolumn{2}{c}{VTAB B0-Inc10} \\ \cline{2-11} 
     & TIA   & Avg   & TIA   & Avg   & TIA   & Avg   & TIA            & Avg            & TIA   & Avg   \\ \hline
EASE & 88.12 & 92.12 & 86.47 & 91.04 & 57.99 & 65.33 & 77.98          & 81.74          & 85.32 & 92.18 \\ \hline
\#1  & 88.45 & 92.46 & 86.59 & 91.13 & 57.93 & 65.98 & 76.25          & 80.46          & 76.73 & 89.00 \\
\#2  & 88.76 & 92.82 & 87.02 & 91.20 & 59.32 & 67.16 & 76.70          & 81.07          & 81.72 & 91.64 \\
\#3  & 89.14 & 92.87 & 87.06 & 91.30 & 60.43 & 67.80 & 78.42 & 81.93 & 85.22 & 92.78 \\
MoTE &
  \textbf{89.43} &
  \textbf{93.06} &
  \textbf{87.66} &
  \textbf{91.90} &
  \textbf{61.23} &
  \textbf{68.44} &
  \textbf{78.50} &
  \textbf{82.07} &
  \textbf{85.90} &
  \textbf{93.16} \\ \hline
\end{tabular}
}
\end{table}

Since class-incremental learning is conducted under task-agnostic settings, the accuracy of task identification in a multi-expert framework can also serve as a meaningful indicator of overall model performance~\cite{EASE,Hide}. In this subsection, we empirically validate the effectiveness of expert filtering and expert joint inference strategies for task ID recognition. As shown in Tab.~\ref{tab:TaskidCompare}, we evaluate the task ID accuracy (\textbf{TIA}) of different model configurations on the test samples and compare our approach with the current SOTA method, EASE. The results indicate that \textbf{(\#2)} incorporating the expert filtering strategy leads to a noticeable improvement in TIA compared to \textbf{(\#1)}. Furthermore, \textbf{(\#3)} integrating expert joint inference further enhances TIA, as this strategy (SCS) emphasizes key experts during joint inference, which is generally effective. Consequently, the full MoTE model achieves an improvement over \textbf{(\#3)}.

\subsubsection{Comparison of Two Adapter Embedding Types}
\label{sec:adapterCompare}
\begin{table}[!t]
\caption{Performance comparison of two different Adapter embedding methods. Bold text indicates superior performance.}
\label{tab:adapterCompare}
\centering
\vspace{-2mm}
\resizebox{\textwidth}{!}{
\begin{tabular}{ccccccccccc}
\hline
\multirow{2}{*}{Type} &
  \multicolumn{2}{c}{CIFAR100 B0-Inc10} &
  \multicolumn{2}{c}{CUB B0-Inc20} &
  \multicolumn{2}{c}{INA B0-Inc10} &
  \multicolumn{2}{c}{INR B0-Inc20} &
  \multicolumn{2}{c}{VTAB B0-Inc10} \\ \cline{2-11} 
    & Last  & Avg   & Last           & Avg            & Last  & Avg   & Last  & Avg   & Last  & Avg   \\ \hline
Seq & 88.76 & 93.05 & \textbf{86.94} & \textbf{91.87} & 56.42 & 67.30 & 75.68 & 81.51 & 84.13 & 92.32 \\
Par &
  \textbf{88.92} &
  \textbf{93.06} &
  86.77 &
  91.83 &
  \textbf{57.74} &
  \textbf{68.43} &
  \textbf{75.93} &
  \textbf{81.78} &
  \textbf{84.78} &
  \textbf{93.16} \\ \hline
\end{tabular}
}
\end{table}

In this subsection, we provide a detailed comparison of the impact of two different adapter embedding methods on algorithm performance. As shown in Tab.~\ref{tab:adapterCompare}, both embedding methods are competitive; however, the "parallel" embedding method achieves superior performance across most benchmarks. Consequently, in our benchmark comparison experiments, we use the "parallel" embedding method as the default approach.

\subsection{Futher Analysis}
\subsubsection{Comparison of Global and Adaptive Scaling Factors}
\label{sec:gamma}
\begin{figure}[!t]
	\centering
	\begin{subfigure}{0.32\linewidth}
		\includegraphics[width=1\columnwidth]{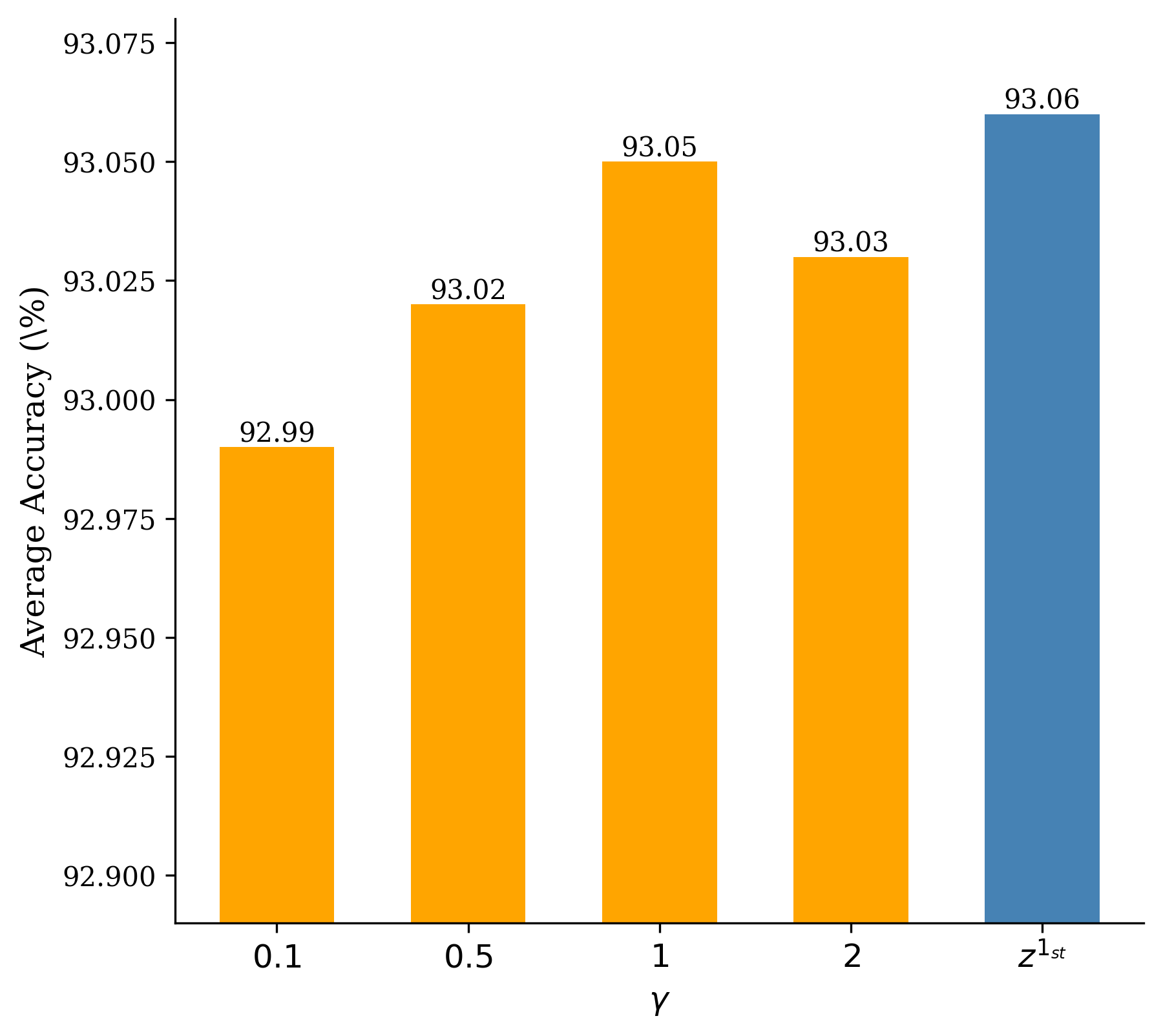}
            \vspace{-7mm}
		\caption{\small CIFAR B0 Inc10}
		\label{fig:gamma-cifar-10}
	\end{subfigure}
	\hfill
	\begin{subfigure}{0.32\linewidth}
		\includegraphics[width=1\linewidth]{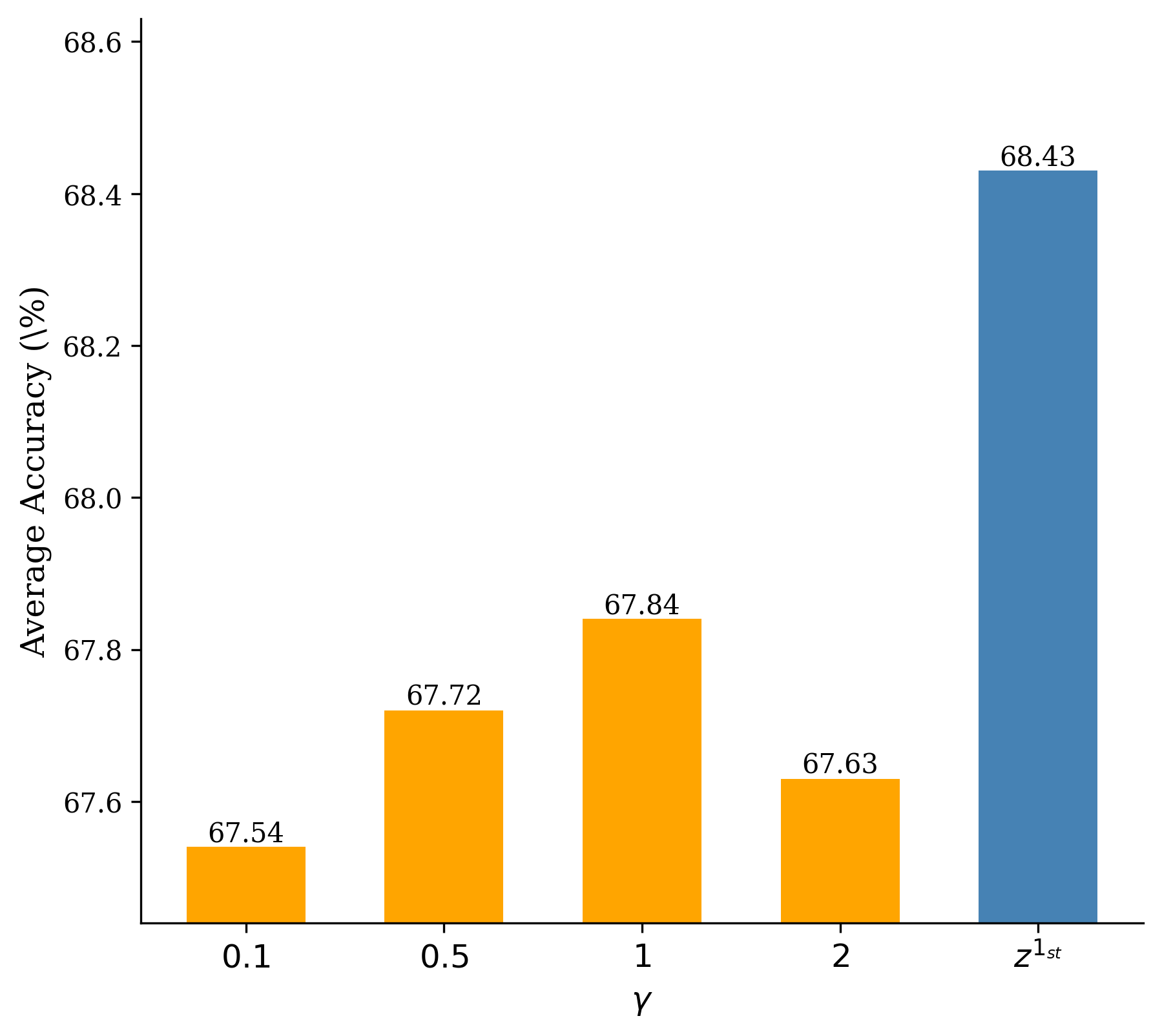}
            \vspace{-7mm}
		\caption{\small INA B0 Inc20}
		\label{fig:gamma-cub-20}
	\end{subfigure}
        \hfill
	\begin{subfigure}{0.32\linewidth}
		\includegraphics[width=1\linewidth]{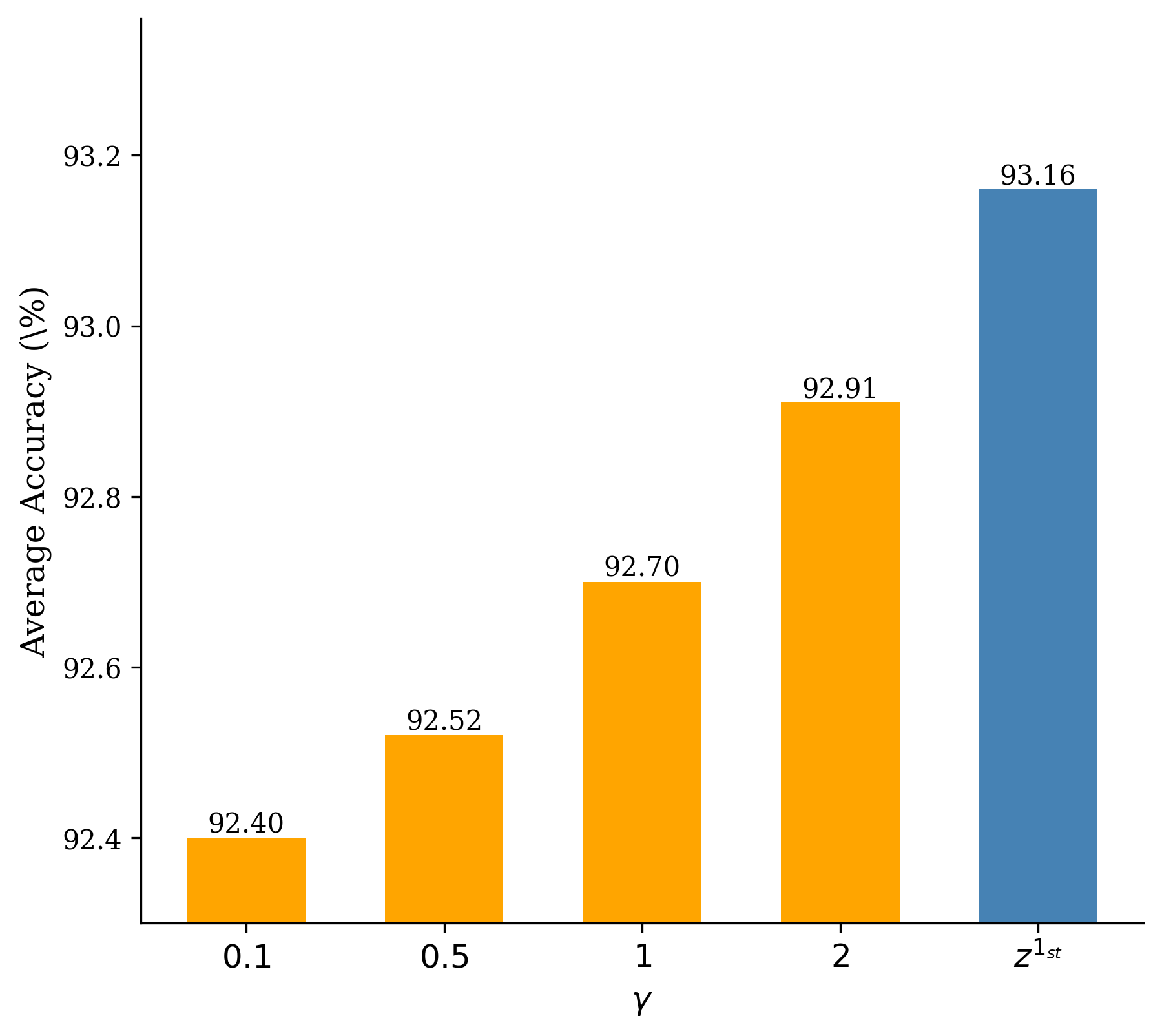}
            \vspace{-7mm}
		\caption{\small VTAB B0 Inc10}
		\label{fig:gamma-INA-20}
	\end{subfigure}
	\vspace{-2mm}
	\caption{\small Performance Comparison of Global vs. Adaptive Scaling, all initialized with (\textbf{ViT-B/16-IN21K})}
	\label{fig:gamma}
\end{figure}

In this subsection, we experimentally investigate the impact of global versus adaptive settings of the scaling factor $\gamma$ in Eq.~\ref{eq:weight} on model performance. We evaluate four fixed values for $\gamma$—{0.1, 0.5, 1, 2}—and compare them against an adaptive setting where $\gamma$ is set to the top confidence score $z^{1^{st}}$. Experiments are conducted on three different benchmark datasets, as shown in Fig.~\ref{fig:gamma}. The results show that the adaptive setting consistently yields the best performance, demonstrating the advantage of adaptive parameterization in enhancing the model's generalization ability.

\subsubsection{Comparison of Inference Time and Memory Costs}
\label{comparison time}
\begin{figure}[!t]
	\centering
	\begin{subfigure}{0.32\linewidth}
		\includegraphics[width=1\columnwidth]{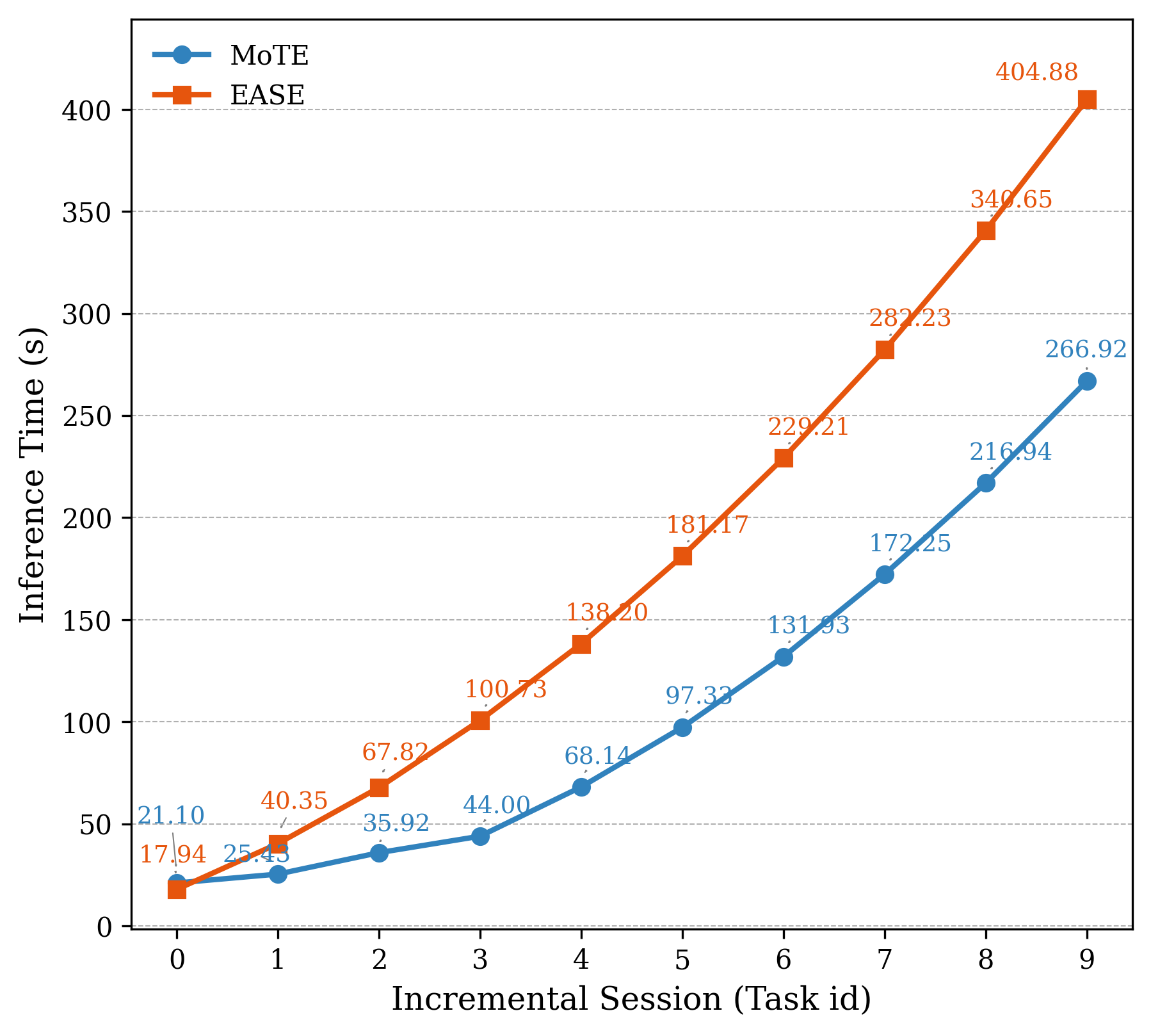}
            \vspace{-7mm}
		\caption{\small CIFAR100 B0 Inc10}
		\label{fig:timegap-cifar-10}
	\end{subfigure}
	\hfill
	\begin{subfigure}{0.32\linewidth}
		\includegraphics[width=1\linewidth]{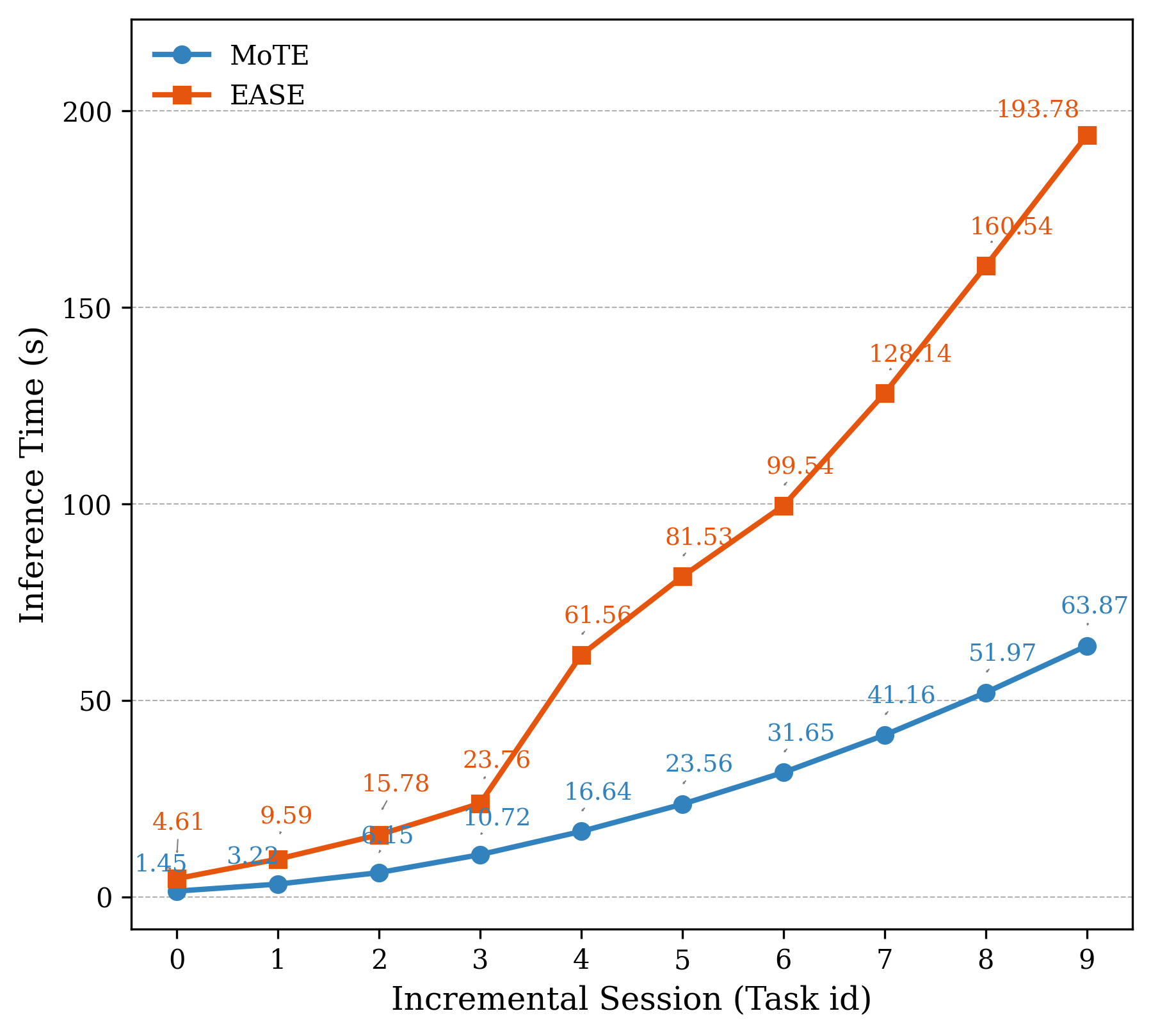}
            \vspace{-7mm}
		\caption{\small CUB B0 Inc20}
		\label{fig:timegap-cub-20}
	\end{subfigure}
        \hfill
	\begin{subfigure}{0.32\linewidth}
		\includegraphics[width=1\linewidth]{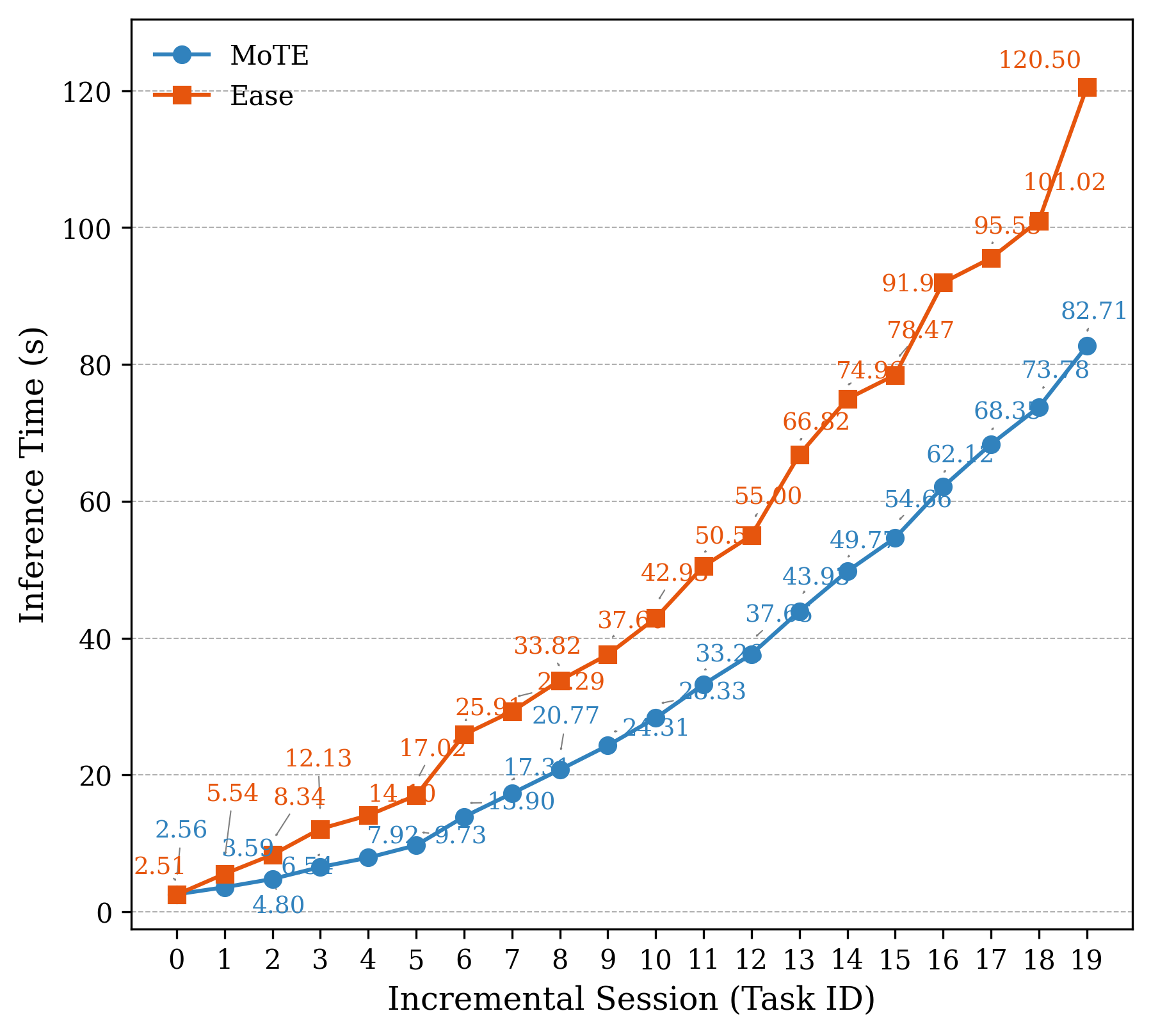}
            \vspace{-7mm}
		\caption{\small INA B0 Inc10}
		\label{fig:timegap-INA-20}
	\end{subfigure}

        \begin{subfigure}{0.32\linewidth}
		\includegraphics[width=1\linewidth]{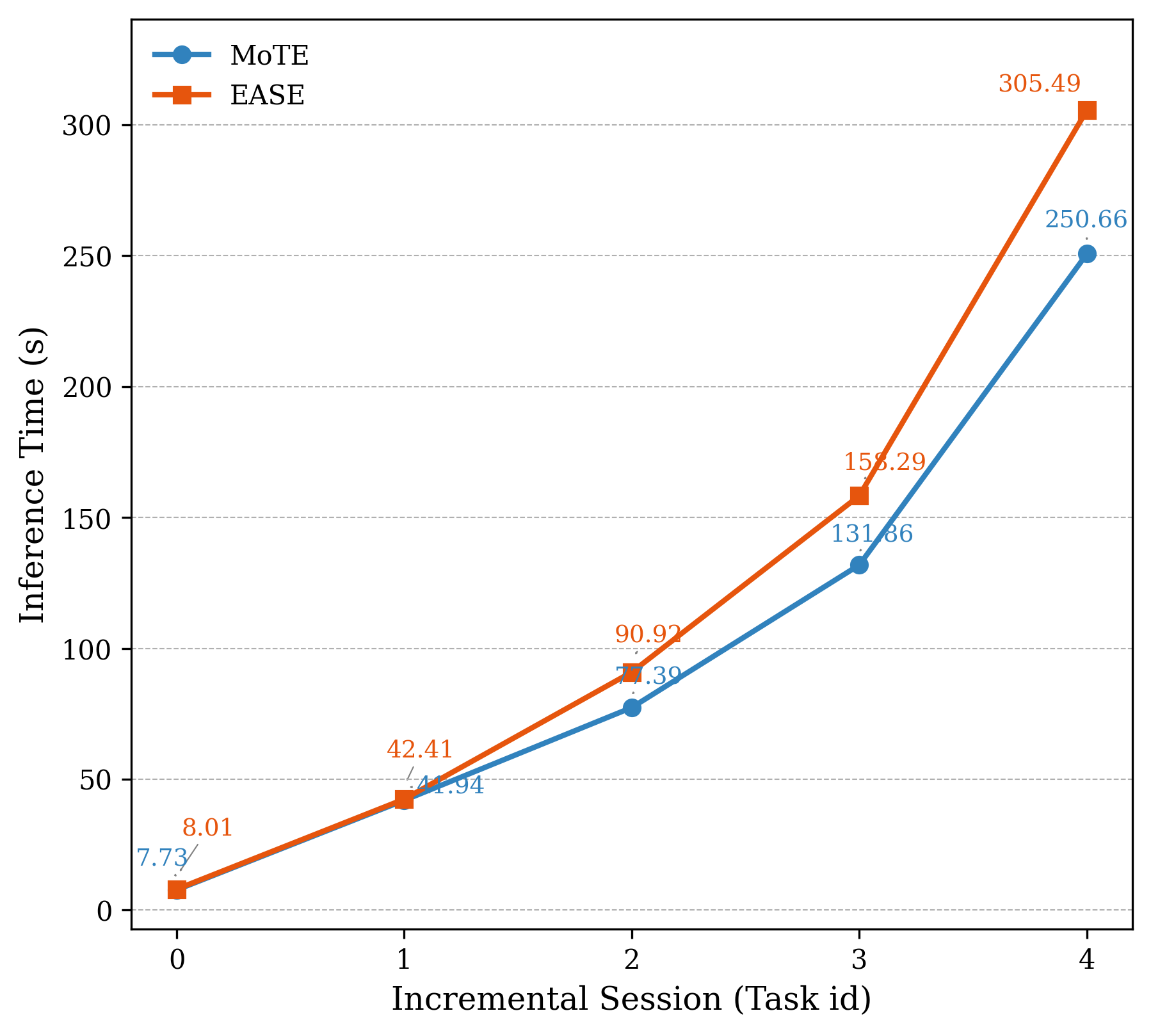}
            \vspace{-7mm}
		\caption{\small VTAB B0 Inc10}
		\label{fig:timegap-VTAB_5}
	\end{subfigure}
        \hfill
        \begin{subfigure}{0.64\linewidth}
		\includegraphics[width=1\linewidth]{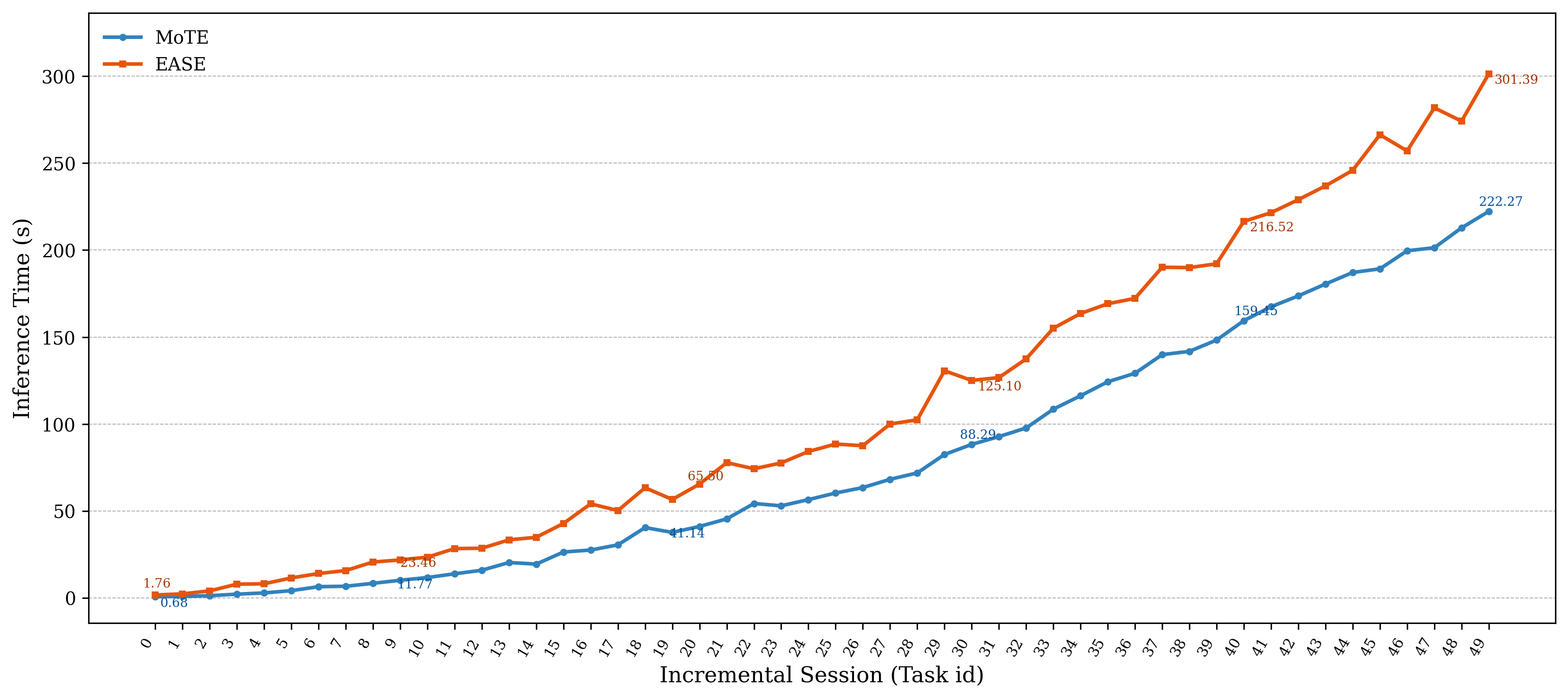}
            \vspace{-7mm}
		\caption{\small INA B0 Inc4}
		\label{fig:timegap-INA_50}
	\end{subfigure}
        
	\vspace{-2mm}
	\caption{\small Comparison of Inference Time with EASE. MoTE and EASE are based on the same PTM ({\bf ViT-B/16-IN21K}).}
	\label{fig:timegap}
\end{figure}

In this subsection, we present a detailed comparison of EASE and MoTE in terms of inference time and memory consumption. As shown in Fig.\ref{fig:timegap}, we evaluate the inference time of both methods across different task settings (i.e., 5, 10, 20, and 50 tasks). It can be observed that MoTE consistently achieves around 30\% faster inference compared to EASE. This improvement mainly stems from MoTE's lightweight design, which eliminates the need to compute pseudo-class prototypes during inference. The architectural differences between the two approaches are also reflected in classifier memory usage. As shown in Tab.~\ref{tab:large}, we report the memory overhead for adapter expansion and class prototype storage in large-scale task scenarios. Since both methods introduce a new adapter per task, the memory usage from adapters is comparable. However, EASE requires additional memory to store pseudo-class prototypes, leading to significantly higher overhead. In contrast, MoTE only stores actual class prototypes, resulting in prototype memory usage approximately $\frac{1}{n}$ of EASE’s (assuming n tasks).

Overall, MoTE not only demonstrates superior inference efficiency and lower memory footprint but also achieves better performance in large-scale class-incremental learning.

\begin{table}[t]
\caption{Comparison of memory consumption and performance between EASE and MoTE in the large-scale task. MoTE and EASE are based on the same PTM (\textbf{ViT-B/16-IN21K}). \textbf{A-MC} denotes the adapter memory costs, and \textbf{P-MC} denotes the prototype memory costs. Both are measured in megabytes (MB).}
\label{tab:large}
\centering
\vspace{-2mm}
\resizebox{\textwidth}{!}{
\begin{tabular}{lccccccccc}
\hline
\multirow{2}{*}{Method} &
  \multicolumn{3}{c}{CIFAR100 B0-Inc2(S50)} &
  \multicolumn{3}{c}{INA B0-Inc4(S50)} &
  \multicolumn{3}{c}{CUB B0-Inc2(S100)} \\ \cline{2-10} 
     & A-MC(MB) & P-MC(MB) & Avg(\%) & A-MC(MB) & P-MC(MB) & Avg(\%) & A-MC(MB) & P-MC(MB) & Avg(\%) \\ \hline
EASE & 225.00      & 14.65   & 84.33   & 225.00      & 29.30   & 54.50     & 450.00      & 29.30   & 91.46   \\
MoTE & 225.00      & \textbf{0.29}    & \textbf{85.94}   & 225.00      & \textbf{0.58}    & \textbf{57.71}     & 450.00      & \textbf{0.58}    & \textbf{92.57}       \\ \hline

\end{tabular}
}
\end{table}

\subsubsection{Visualizations}
\label{Visualizations}
\begin{figure}[!t]
    \centering
	\begin{subfigure}{0.32\linewidth}
		\includegraphics[width=1\columnwidth]{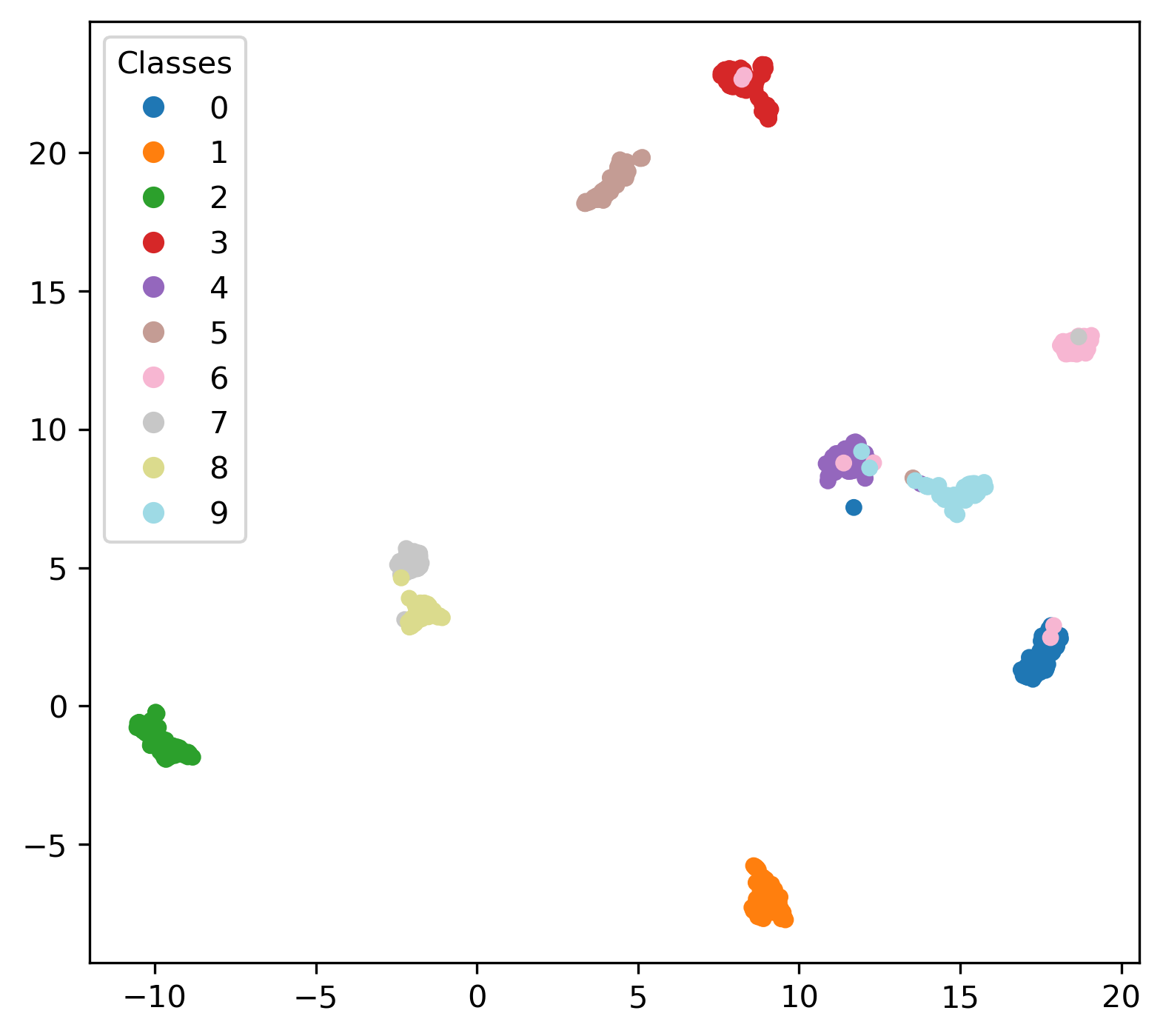}
            \vspace{-7mm}
		\caption{\small Extracted by $A_{1}$}
		\label{fig:tsne-cifar-0}
	\end{subfigure}
	\hfill
	\begin{subfigure}{0.32\linewidth}
		\includegraphics[width=1\linewidth]{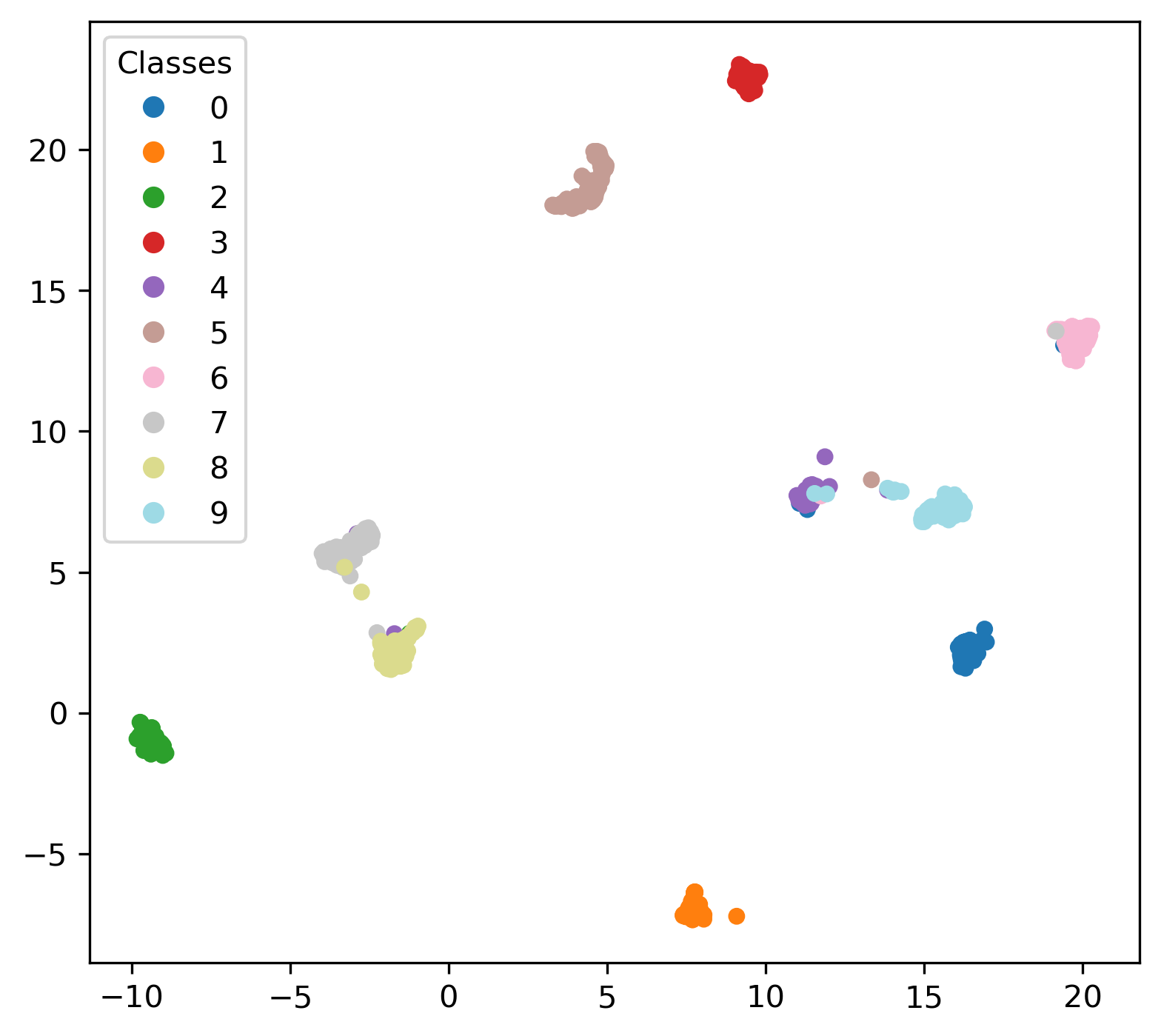}
            \vspace{-7mm}
		\caption{\small Extracted by $A_{2}$}
		\label{fig:tsne-cifar-1}
	\end{subfigure}
        \hfill
	\begin{subfigure}{0.32\linewidth}
		\includegraphics[width=1\linewidth]{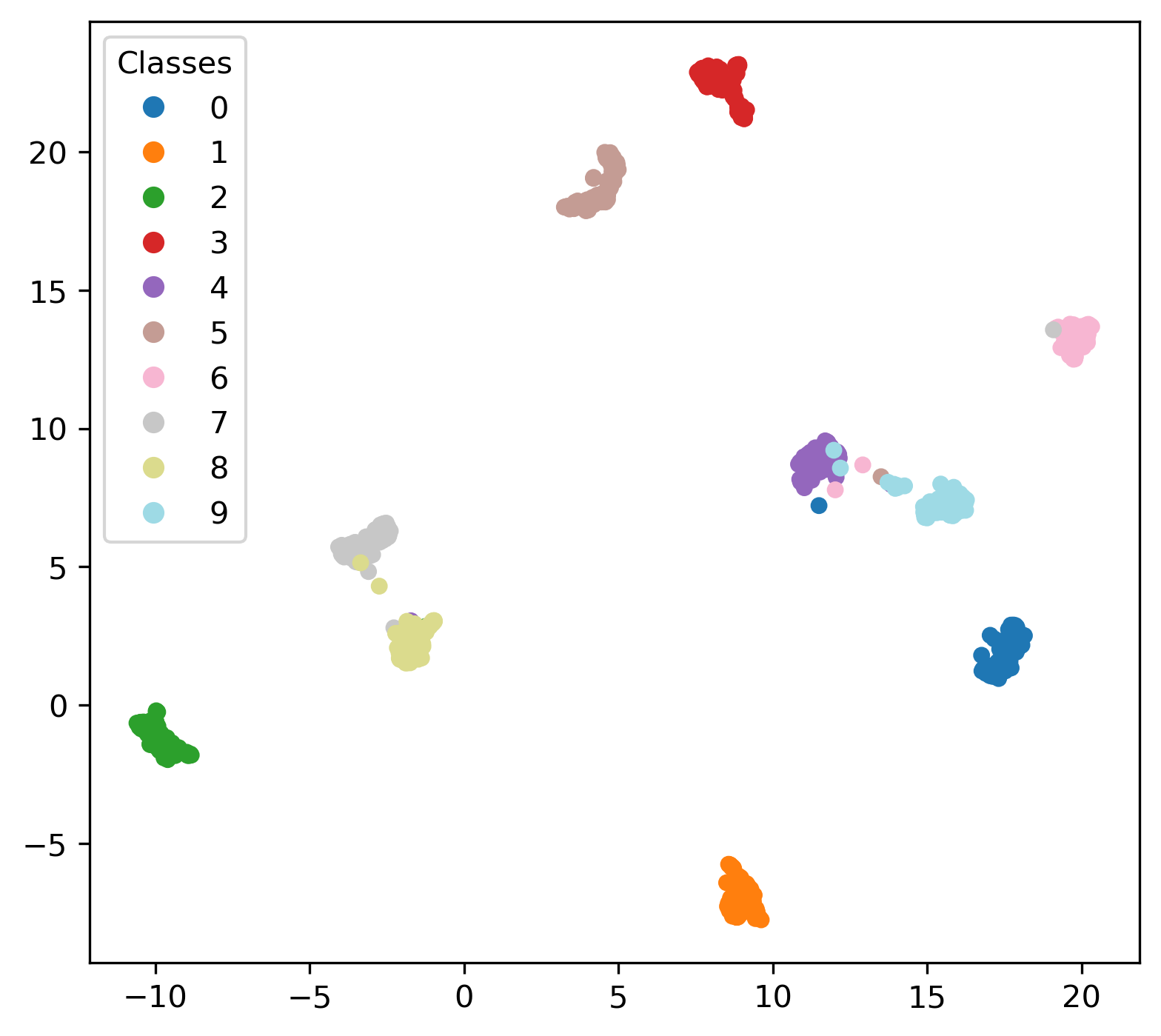}
            \vspace{-7mm}
		\caption{\small Extracted by MoTE}
		\label{fig:tsne-cifar-model}
	\end{subfigure}
    
	\begin{subfigure}{0.32\linewidth}
		\includegraphics[width=1\columnwidth]{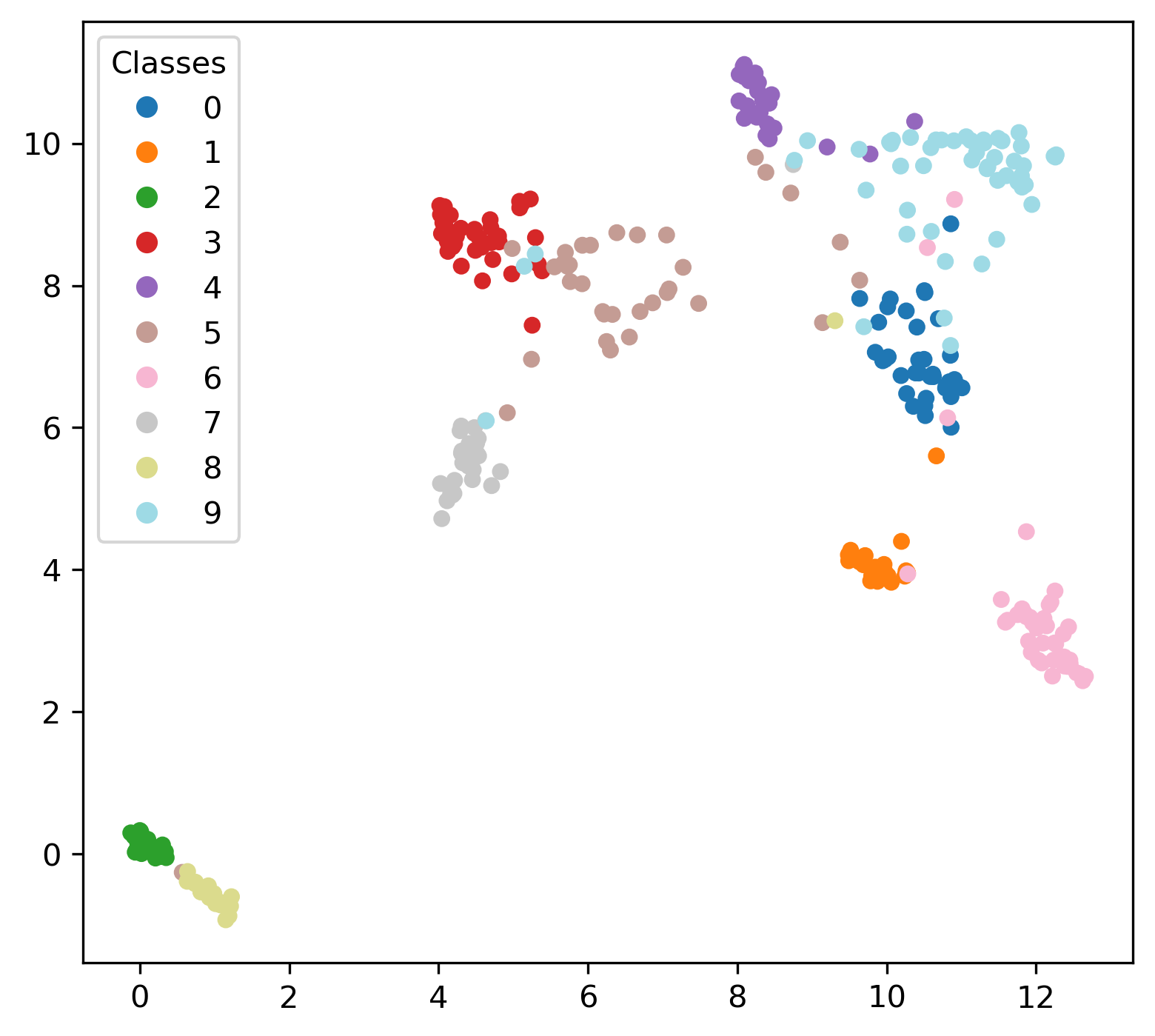}
            \vspace{-7mm}
		\caption{\small Extracted by $A_{1}$}
		\label{fig:tsne-inr-0}
	\end{subfigure}
	\hfill
	\begin{subfigure}{0.32\linewidth}
		\includegraphics[width=1\linewidth]{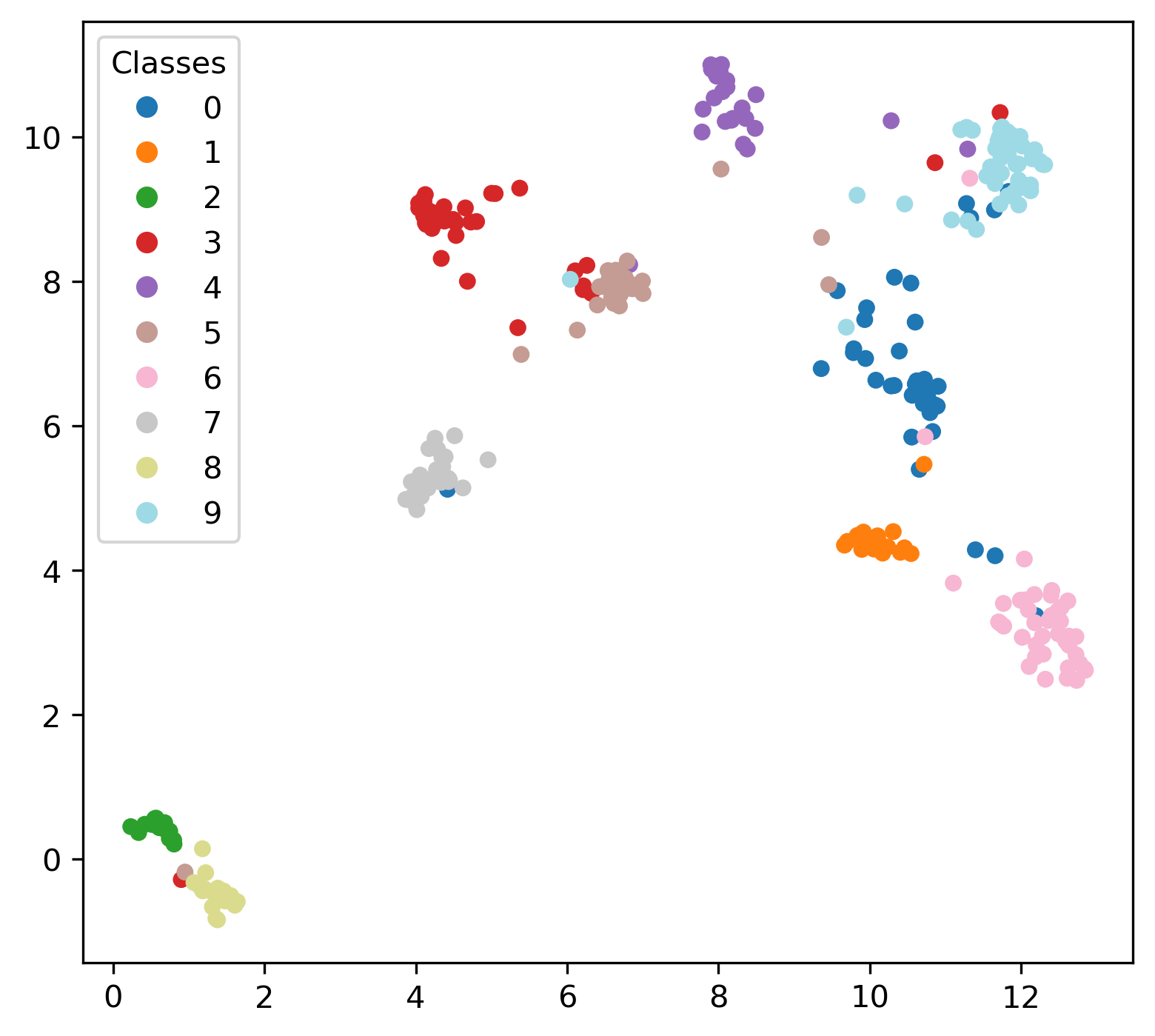}
            \vspace{-7mm}
		\caption{\small Extracted by $A_{2}$}
		\label{fig:tsne-inr-1}
	\end{subfigure}
        \hfill
	\begin{subfigure}{0.32\linewidth}
		\includegraphics[width=1\linewidth]{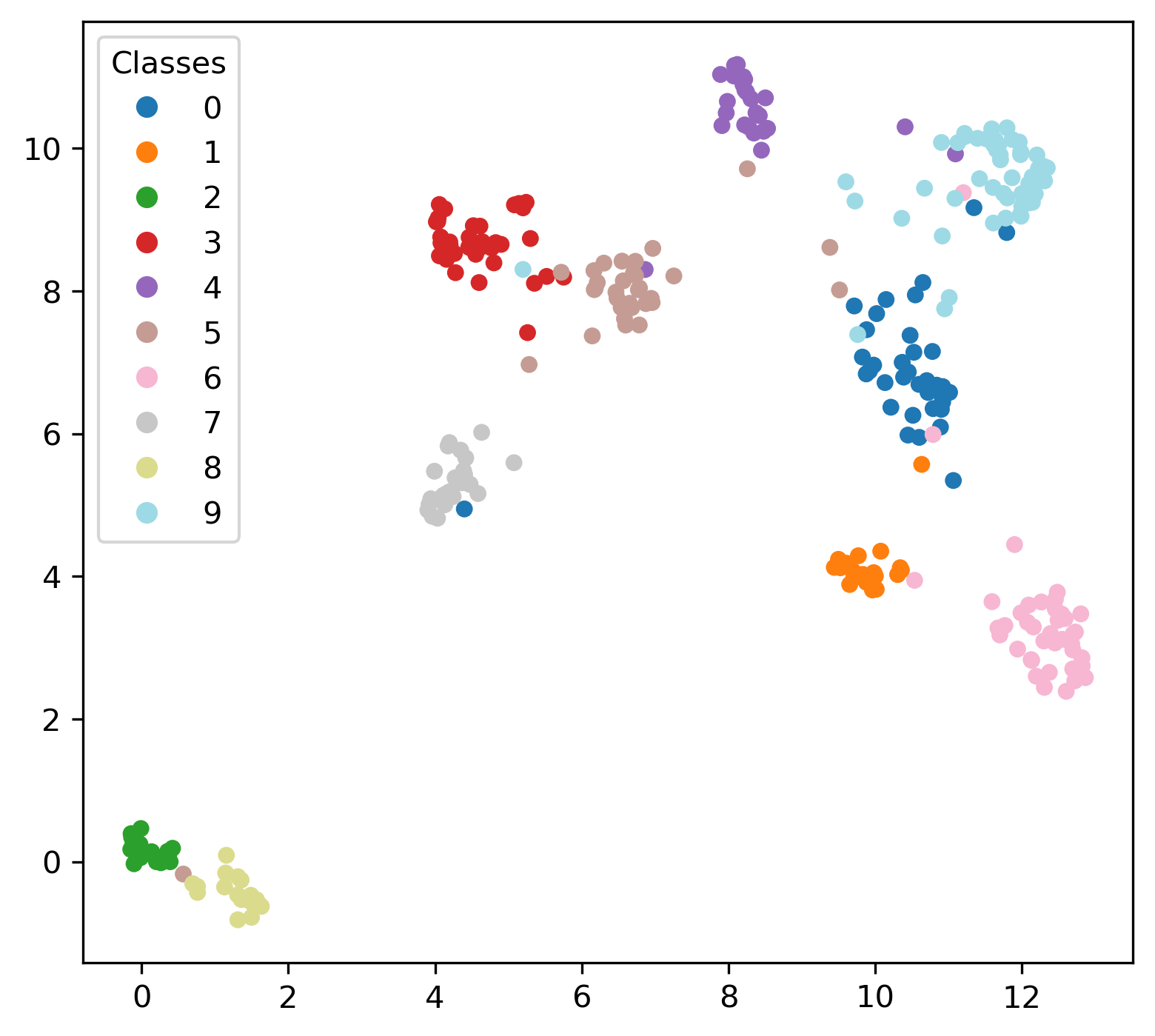}
            \vspace{-7mm}
		\caption{\small Extracted by MoTE}
		\label{fig:tsne-inr-model}
	\end{subfigure}

        \begin{subfigure}{0.32\linewidth}
		\includegraphics[width=1\columnwidth]{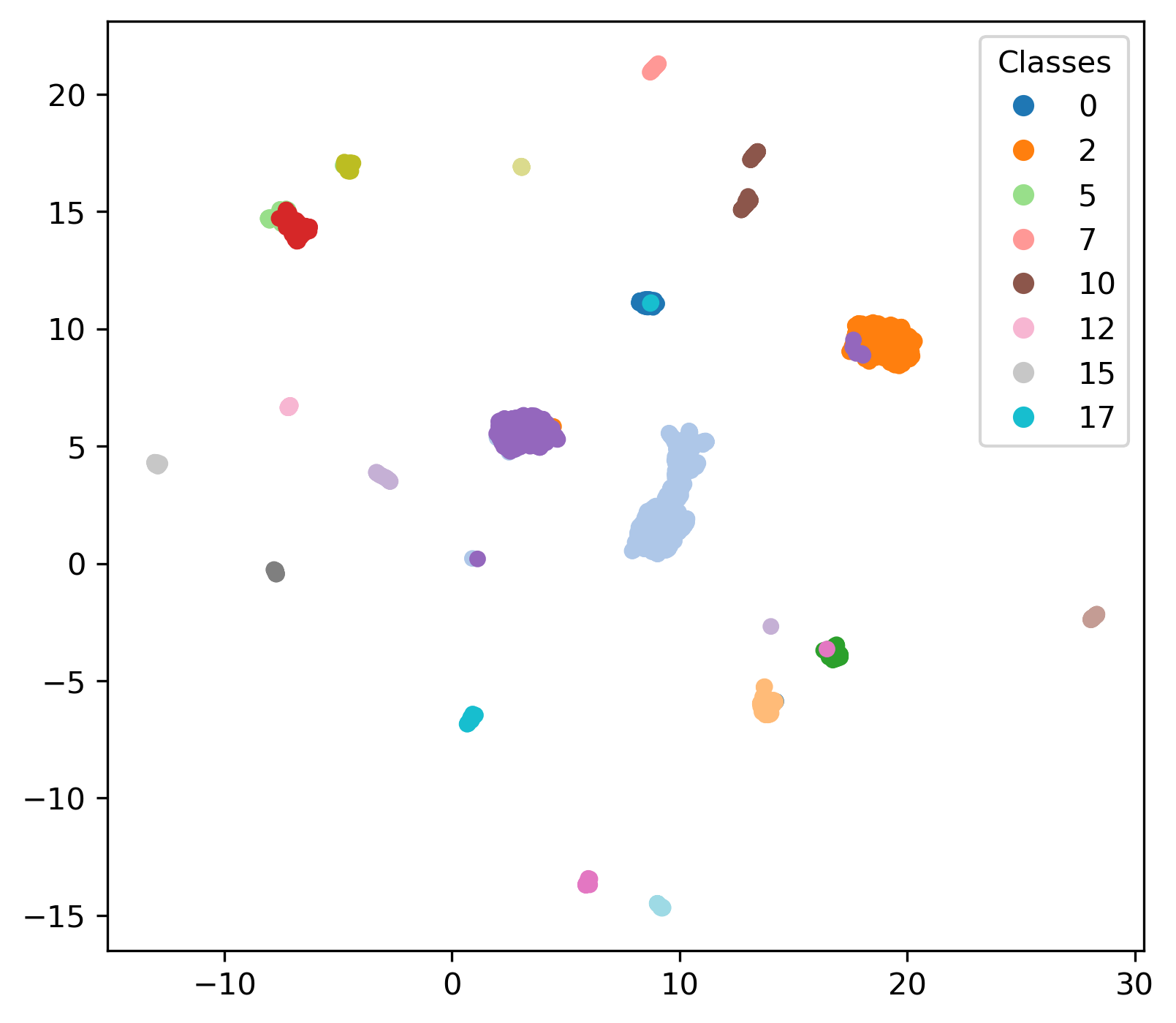}
            \vspace{-7mm}
		\caption{\small Extracted by $A_{1}$}
		\label{fig:tsne-vtab-0}
	\end{subfigure}
	\hfill
	\begin{subfigure}{0.32\linewidth}
		\includegraphics[width=1\linewidth]{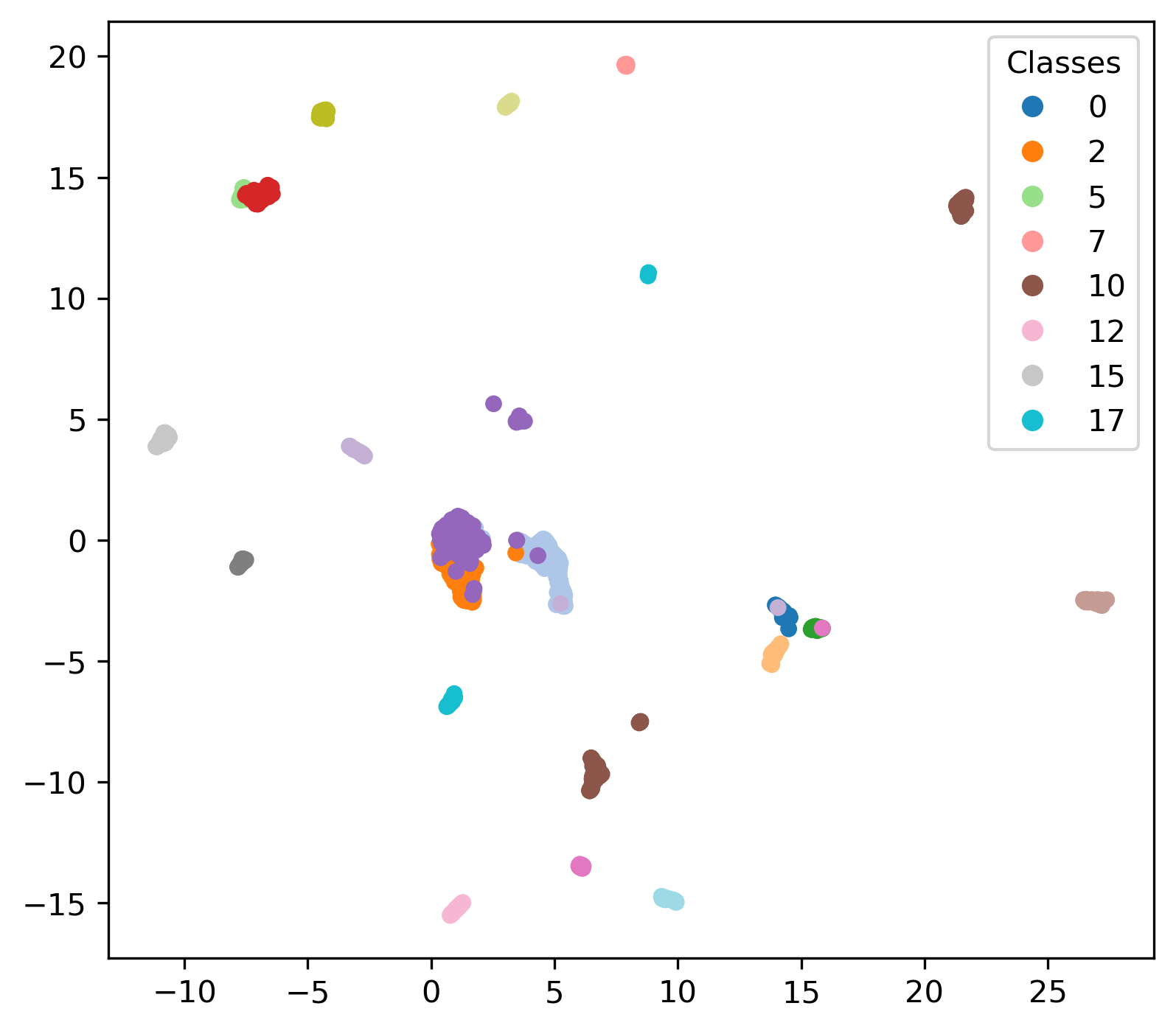}
            \vspace{-7mm}
		\caption{\small Extracted by $A_{2}$}
		\label{fig:tsne-vtab-1}
	\end{subfigure}
        \hfill
	\begin{subfigure}{0.32\linewidth}
		\includegraphics[width=1\linewidth]{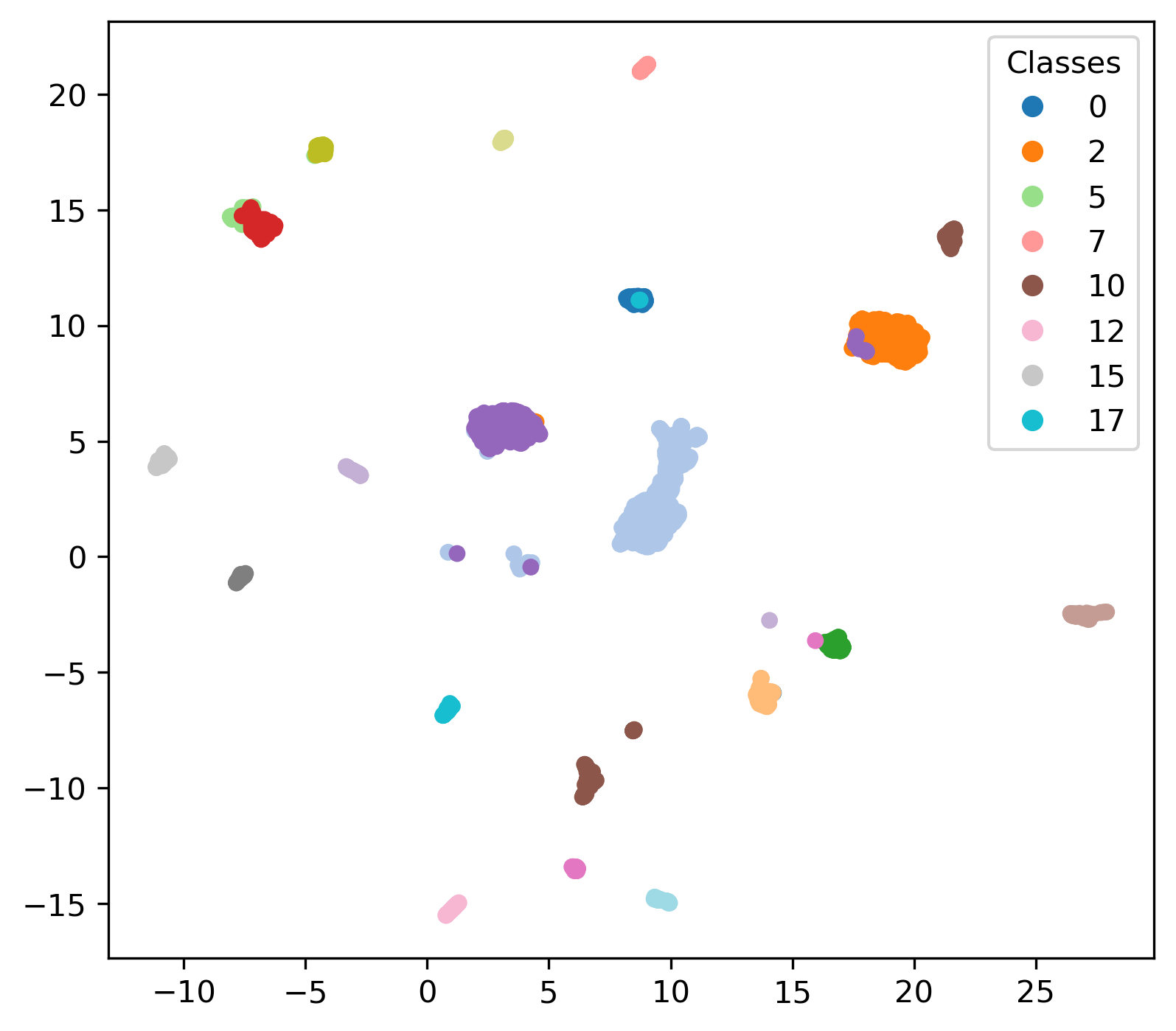}
            \vspace{-7mm}
		\caption{\small Extracted by MoTE}
		\label{fig:tsne-vtab-model}
	\end{subfigure}
        
	\vspace{-2mm}
	\caption{\small Visualization of t-SNE~\cite{TSNE} for feature representations from different experts. (a), (b) and (c) correspond to the experimental protocol CIFAR B0-Inc5, (d), (e), and (f) correspond to the INR B0-Inc5, while (g), (h), and (i) correspond to the VTAB B0-Inc10.}
	\label{fig:Visualization}
\end{figure}

In this paper, we propose a Weighted Multi-Expert Inference mechanism to address the classification challenge of task-ambiguous samples. To validate the effectiveness of this mechanism, we conduct experiments on CIFAR B0-Inc5, ImageNet-R B0-Inc5, and VTAB B0-Inc5 and visualize the embedding distributions of both individual experts and fused experts using t-SNE, as shown in Fig.~\ref{fig:Visualization}. Each incremental stage consists of five new classes, and for each task, we train two adapters, $A_{1}$ and $A_{2}$.

We observe that in the CIFAR B0-Inc5 setting (Fig. 10(a)), $A_{1}$ performs well on Task 1 (classes 0–4), but some samples from classes 5–9 tend to be confused in the feature space. Conversely,  $A_{2}$ excels on classes 5–9 but occasionally misclassifies samples from classes 0–4. While individual experts exhibit a basic ability to handle global classification, their discriminative power diminishes for samples outside their designated task scope, making it challenging to correctly classify task-ambiguous samples. After applying our Weighted Multi-Expert Inference mechanism, the classification performance in the shared feature space is notably improved compared to relying on any single expert. Similar trends are observed on the other two datasets. In ImageNet-R B0-Inc5 (Figs. 10(d), (e), and (f)) and VTAB B0-Inc10 (Figs. 10(g), (h), and (i)), each expert effectively separates features within its task but struggles with out-of-task samples. The proposed fusion strategy consistently mitigates these issues and enhances feature discrimination across tasks. This improvement is further verified through ablation studies presented in Tab.~\ref{tab:abla} and Tab.~\ref{tab:TaskidCompare}.

\subsubsection{Evaluation on Adapter-limited MoTE}
\label{Sec:ablation_limit}
\begin{figure}[!t]
	\begin{subfigure}{0.48\linewidth}
		\includegraphics[width=1\columnwidth]{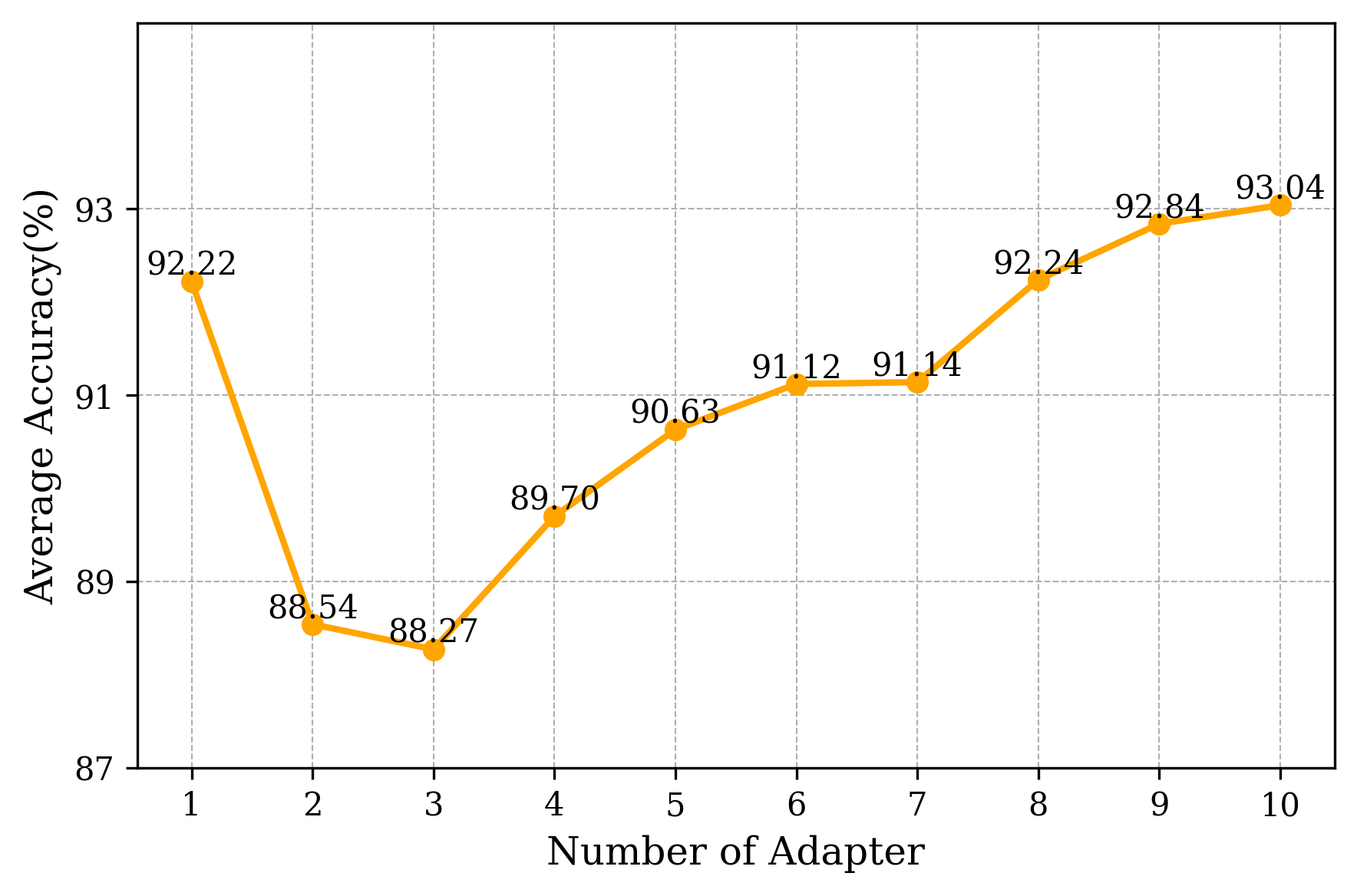}
            \vspace{-7mm}
		\caption{CIFAR100 B0-Inc10}
		\label{fig:CIFAR_limit}
	\end{subfigure}
	\hfill
	\begin{subfigure}{0.48\linewidth}
		\includegraphics[width=1\linewidth]{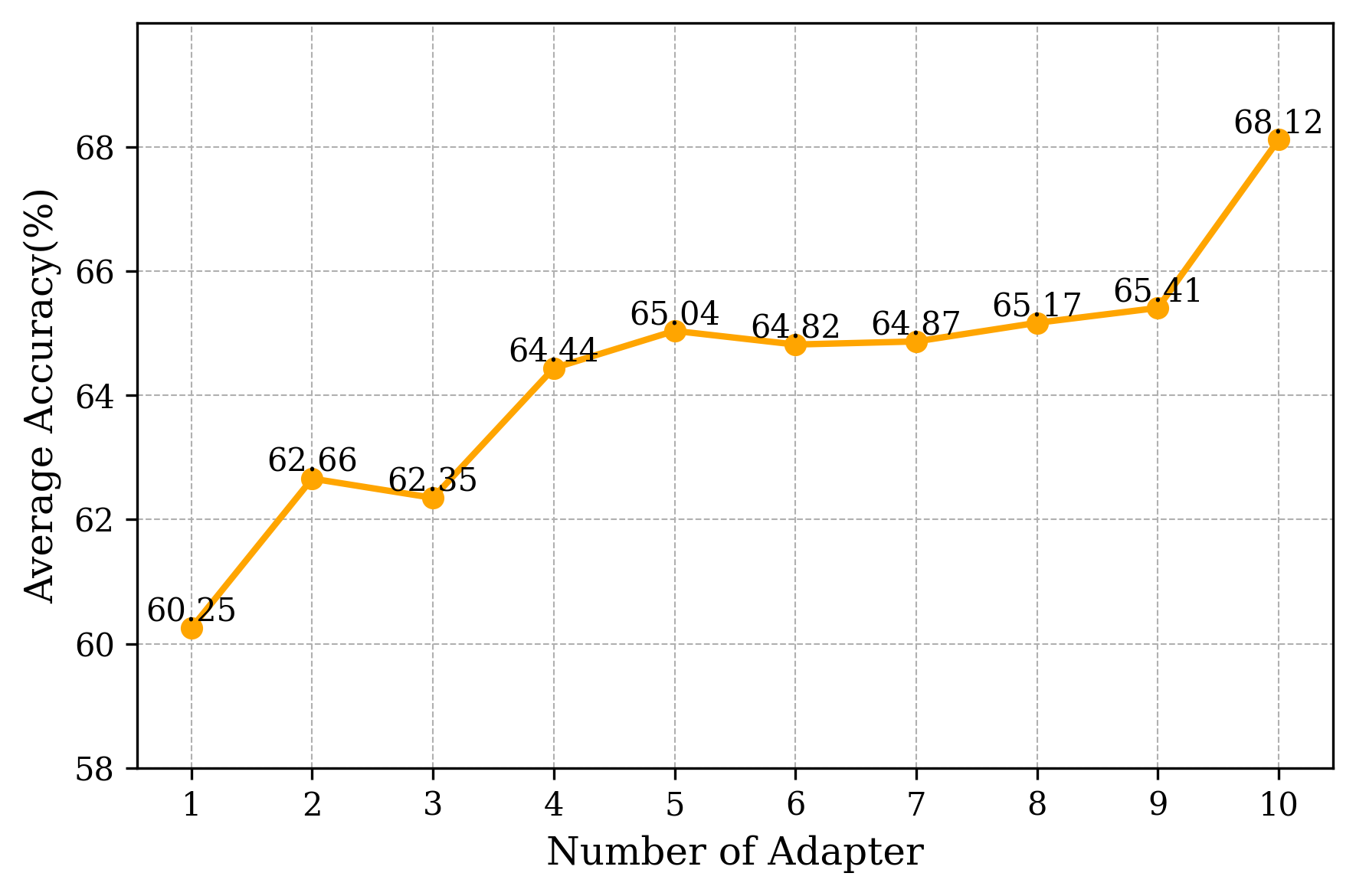}
            \vspace{-7mm}   
		\caption{INA B0-Inc20}
		\label{fig:INA_limit}
	\end{subfigure}
       
        \begin{subfigure}{0.48\linewidth}
		\includegraphics[width=1\linewidth]{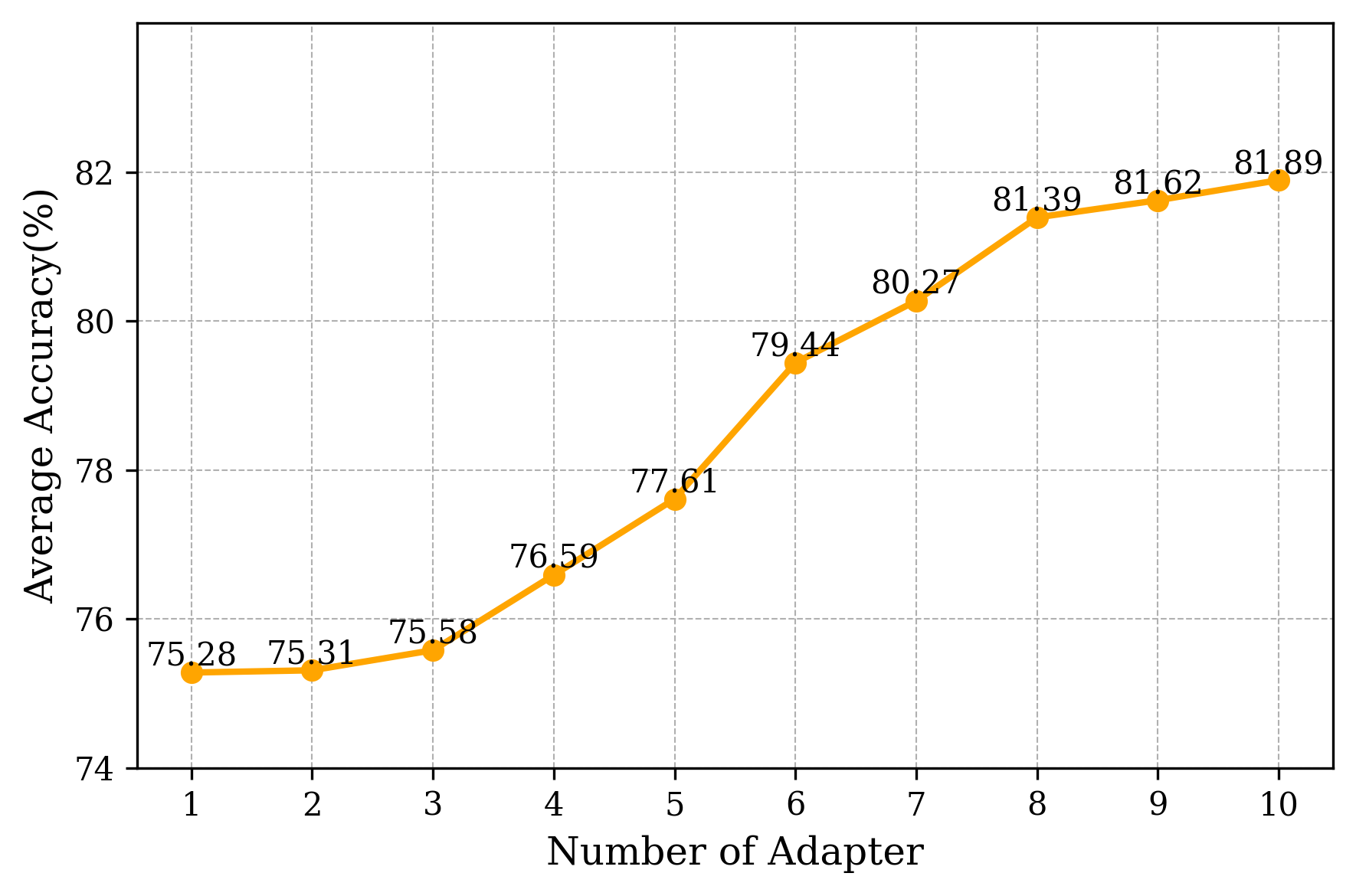}
            \vspace{-7mm}
		\caption{INR B0-Inc20}
		\label{fig:INR_limit}
	\end{subfigure}
         \hfill
        \begin{subfigure}{0.48\linewidth}
		\includegraphics[width=1\linewidth]{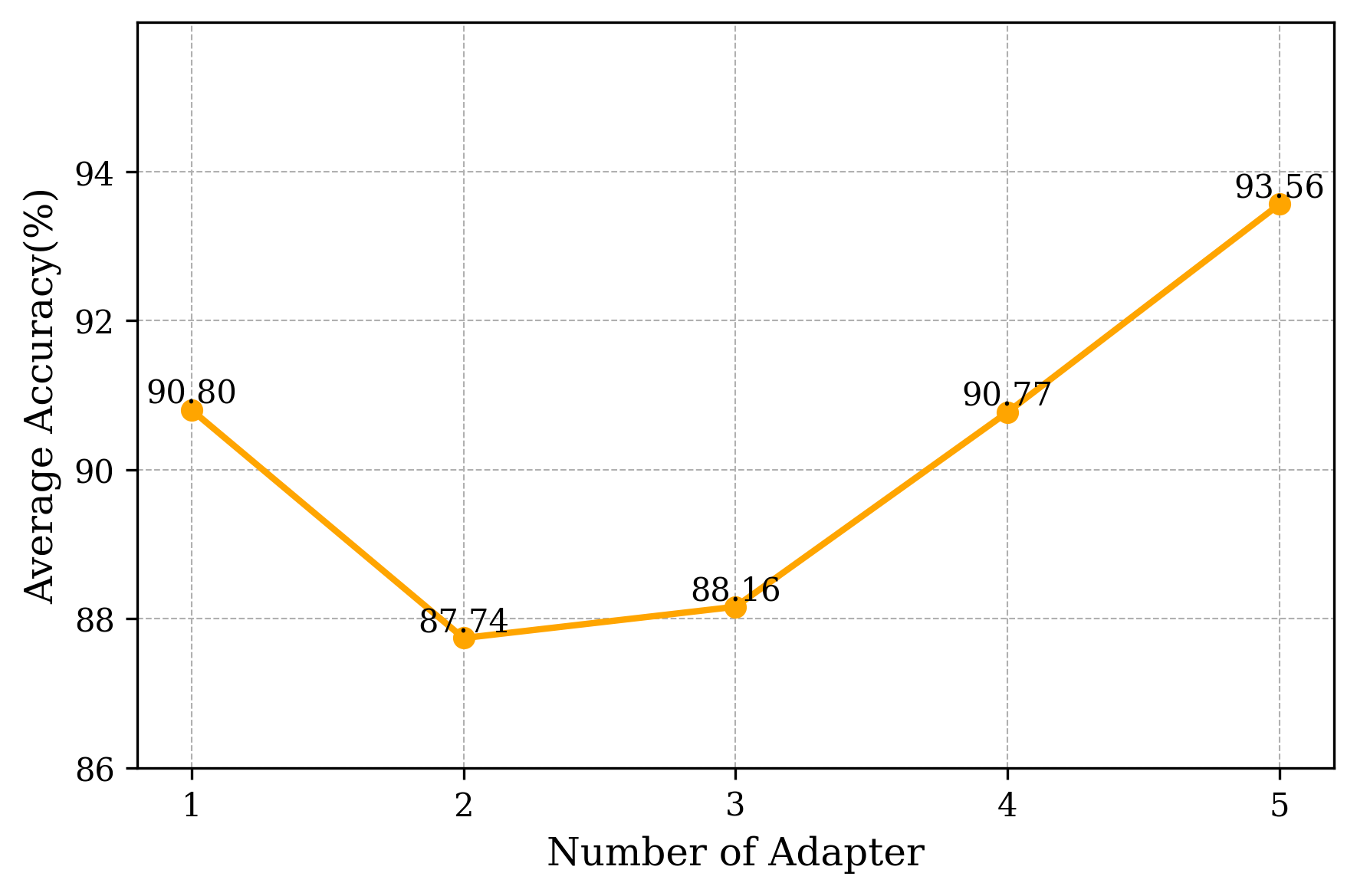}
            \vspace{-7mm}
		\caption{VTAB B0-Inc10}
		\label{fig:VTAB_limit}
	\end{subfigure}
        \vspace{-2mm}
	\caption{\small Performance comparison of different numbers of adapters under various experimental settings, all initialized with \textbf{ViT-B/16-IN21K}.}
	\label{fig:mote_limit}
\end{figure}

We comprehensively evaluated the model’s performance across incremental tasks using four datasets, examining the impact of varying adapter constraints. Our primary objective was to investigate the relationship between the number of adapters and task performance. 

The experimental results across the four datasets indicate that performance improves as more experts are incorporated into the model using the expert evaluation mechanism. Notably, assigning one expert per task yields the best overall performance. However, our findings also reveal that using only a single adapter is not significantly disadvantageous; in certain cases, it surpasses configurations with multiple adapters. This observation suggests that a pre-trained model inherently retains a baseline zero-shot discrimination capability for subsequent downstream tasks when adapted to a single task via an adapter.

As illustrated in Figs.\ref{fig:CIFAR_limit} and \ref{fig:VTAB_limit}, for the CIFAR-100 and VTAB datasets, a single adapter is sufficient for effective adaptation to downstream tasks. Interestingly, introducing multiple adapters initially degrades performance, and only after reaching a critical threshold does performance recover to the level achieved with a single adapter. This phenomenon can be attributed to the relative simplicity of CIFAR-100 and VTAB, where inter-task feature variations are minimal. Training a single adapter on the first task enables the model to generalize to subsequent tasks to some extent. However, when multiple adapters are introduced, the adapter-limited MoTE framework requires them to collaborate in generating class prototypes for unknown tasks. In the case of simple downstream tasks, this joint representation introduces noise due to the limited number of available adapters, thereby impairing performance relative to the single-adapter setting. As the number of adapters increases beyond a certain threshold, the growing pool of task-specific experts reduces reliance on joint inference, mitigating noise effects and ultimately surpassing the performance of the single-adapter configuration. In contrast, Figures \ref{fig:INA_limit} and \ref{fig:INR_limit} illustrate results on the more complex ImageNet-A and ImageNet-R datasets, where inter-task differences are more pronounced. In these cases, progressively increasing the number of adapters consistently enhances performance, as the expansion of task-specific experts enables better adaptation to the diverse task distributions.

While assigning one adapter per task yielded optimal performance, the observed improvement was relatively marginal. This finding suggests that the relationship between the number of experts and task performance is not strictly linear. We hypothesize that refining the granularity of expert allocation from the task level to the feature level could further optimize this trade-off, potentially leading to improved model adaptability and efficiency.

\section{Conclusion}
Class incremental learning requires models to continuously absorb new knowledge from data streams while retaining previously learned information. The advancement of pre-trained models has accelerated the practical application of Class incremental learning. In this paper, we propose the mixture of task-specific expert (MoTE) framework aimed at producing robust mixed features and enhancing the classification capabilities of pre-trained models, even when task boundaries are unclear. Specifically, we train a lightweight adapter for each task to capture task-specific information, allowing it to act as a dedicated task expert. During inference, we filter out unreliable ones and select the most reliable expert at an instance level. The final prediction is obtained by a weighted fusion of the selected experts' outputs. Extensive experiments demonstrate the effectiveness of MoTE, with an inference speed nearly 30$\%$ faster than comparable methods. Furthermore, to explore the relationship between the number of adapters and the number of tasks, we designed Adapter-Limited MoTE based on MoTE and conducted extensive experiments to investigate this relationship. We hope that our findings will provide insights and inspiration for future research.

\textbf{Future work:} In MoTE, an adapter is added and trained independently for each task. Although the adapter has a small parameter count(0.3$\%$ of the total backbone), the increase in the number of tasks can lead to memory and storage overhead. Our current approach involves training task-specific adapters, but future work could explore feature-specific adapters that adapt based on the feature distribution at different layers. This adaptive expansion strategy could achieve state-of-the-art performance while minimizing storage costs. Additionally, this study focuses on the class-incremental learning paradigm, which inherently assumes certain conditions, such as non-overlapping classes between tasks, each sample being assigned a single label, and a consistent domain across all samples. Exploring how CIL can be integrated with other techniques to address these limitations is crucial for advancing toward more generalized incremental learning. While MoTE is designed for standard CIL, its core components (task-specific adapters, prototype-based filtering) provide a foundation for addressing more complex settings. To further bridge the gap between class-incremental learning (CIL) settings and real-world applications, we go beyond conventional CIL and explore a more realistic scenario: \textbf{cross-dataset class-incremental learning}. Traditional CIL assumes disjoint class distributions across tasks and single-label samples, and most existing benchmarks simulate task sequences by splitting a single dataset—for example, CIFAR100 B0-Inc10—meaning that all samples during training and inference come from the same data domain. However, in real-world scenarios, the incoming data stream may originate from multiple datasets with different styles or domains. To better simulate such realistic settings, we merge multiple datasets and construct two cross-dataset class-incremental experiments: \textbf{CIFAR100-CUB-INR} and \textbf{INR-CUB-CIFAR100 (B0-Inc20)}. We report the average accuracy after completing each dataset in the sequence, as shown in Tab.~\ref{tab:Multidatasets}. The results demonstrate that our proposed MoTE still achieves strong performance under these challenging conditions, indicating good generalizability. It is worth noting that MoTE was not specifically designed for handling domain shifts across tasks. We hope this preliminary exploration can inspire future research toward more general and robust continual learning algorithms.

% \usepackage{multirow}
\begin{table}[t]
\caption{Performance comparison of cross-dataset CIL. 'Avg-\textit{n}' denotes the average accuracy after completing training on each individual dataset. All experiments are conducted
using the same pre-trained model (\textbf{ViT-B/16-IN21K}) and \textbf{B0-Inc20}. Bolded values indicate the best performance.}
\vspace{-2mm}
\label{tab:Multidatasets}
\resizebox{\textwidth}{!}{
\begin{tabular}{l|ccc|ccc}
\hline
\multirow{2}{*}{Method} & \multicolumn{3}{c|}{CIFAR100$\rightarrow$CUB$\rightarrow$INR}   & \multicolumn{3}{c}{INR$\rightarrow$CUB$\rightarrow$CIFAR100}    \\ \cline{2-7} 
              & Avg-1 & Avg-2 & Avg-3 & Avg-1 & Avg-2 & Avg-3 \\ \hline
L2P           & 91.76 & 86.24 & 79.95 & 76.08 & 71.51 & 70.44 \\
DualPrompt    & 91.64 & 86.52 & 80.32 & 73.62 & 70.27 & 69.54 \\
CODA\_Prompt  & 92.29 & 87.15 & 81.27 & 76.97 & 74.32 & 73.27 \\
ADAM\_Adapter & 92.44 & 89.87 & 86.69 & 74.50 & 72.75 & 73.81 \\
EASE          & 93.00 & 90.57 & 87.86 & 81.74 & 77.97 & 78.52 \\ \hline
MoTE                    & \textbf{93.62} & \textbf{91.31} & \textbf{88.40} & \textbf{82.07} & \textbf{78.76} & \textbf{78.93} \\ \hline
\end{tabular}
}
\end{table}

\section*{Acknowledgments}
The paper is supported by the National Natural Science Foundation of China (62101061).

\bibliographystyle{elsarticle-num}
\bibliography{main}

\end{document}